\documentclass[pmlr]{jmlr}

\usepackage{longtable}
\usepackage{booktabs}
\usepackage{hyperref}
\usepackage{url}
\usepackage{amsfonts,amssymb,amsmath,mathtools}
\usepackage{appendix} 
\usepackage{bbm}
\usepackage{lineno}
\usepackage{graphicx,csquotes}
\usepackage{booktabs , multirow}

\jmlrvolume{260}
\jmlryear{2024}
\jmlrworkshop{ACML 2024}
\jmlrpages{-}
\renewcommand{\thepage}{}
\def \x {\mathbf{x}}
\title[]{Towards Calibrated Losses for Adversarial Robust Reject Option Classification}

\author{\Name{Vrund Shah} \Email{vrund.shah@research.iiit.ac.in}\\
\Name{Tejas Chaudhari} \Email{tejas.chaudhari@research.iiit.ac.in}\\
\Name{Naresh Manwani} \Email{naresh.manwani@iiit.ac.in}\\
  \addr Machine Learning Lab, Kohli Research Block, IIIT Hyderabad - 500032, India}

\editors{Vu Nguyen and Hsuan-Tien Lin}

\begin{document}

\maketitle

\begin{abstract}
Robustness towards adversarial attacks is a vital property for classifiers in several applications such as autonomous driving, medical diagnosis, etc. Also, in such scenarios, where the cost of misclassification is very high, knowing when to abstain from prediction becomes crucial. A natural question is which surrogates can be used to ensure learning in scenarios where the input points are adversarially perturbed and the classifier can abstain from prediction? This paper aims to characterize and design surrogates calibrated in \enquote{Adversarial Robust Reject Option} setting. First, we propose an adversarial robust reject option loss $\ell_{d}^{\gamma}$ and analyze it for the hypothesis set of linear classifiers ($\mathcal{H}_{\textrm{lin}}$). Next, we provide a complete characterization result for any surrogate to be $(\ell_{d}^{\gamma},\mathcal{H}_{\textrm{lin}})$- calibrated. To demonstrate the difficulty in designing surrogates to $\ell_{d}^{\gamma}$, we show negative calibration results for convex surrogates and quasi-concave conditional risk cases (these gave positive calibration in adversarial setting without reject option). We also empirically argue that Shifted Double Ramp Loss (DRL) and Shifted Double Sigmoid Loss (DSL) satisfy the calibration conditions. Finally, we demonstrate the robustness of shifted DRL and shifted DSL against adversarial perturbations on a synthetically generated dataset.

\end{abstract}


\section{Introduction}
Many machine learning models are susceptible to adversarial attacks \citep{Goodfellow_Explain_Adv_attk,Szegedy_Intrigue_prop_NN}, i.e., imperceivable changes in the data at testing time results in learning of bad 
classifiers and even hazardous accidents.
For example, the presence of an artifact on a traffic sign may lead to inaccurate interpretation of the signal, often arising in autonomous driving. To address such problems, several studies were conducted for learning classifiers with reduced sensitivity to these perturbations \citep{Aditi_Certified_Defense_Adv_Ex, Wong_Kolter_Adv_Ex_Cvx}. The property displayed by these classifiers is \enquote{Adversarial Robustness}.

To achieve robustness against adversarial attacks, the worst-case loss subject to adversarial perturbations is used. Adversarial robustness to small $l_{p}$-norm perturbations has been analyzed in \citet{Carlini_Wagner_Eval_Rob_NN, Madry_DL_Resist_Adv_Attk}. Optimization of adversarial loss is hard, and this calls for the use of surrogates. An important property that the surrogates should satisfy is \enquote{Consistency} - i.e., minimization of the true risk associated with surrogate loss should lead to the minimization of the true risk associated with target loss. One way to analyze consistency is using \enquote{calibration} - point-wise minimization of the conditional risk \citep{Bartlett_Jordan_Conv_Clssfn_RB, Steinwart_Compare_Diff_Losses}. \citet{Bao_CSL_Adv_Rob} showed that convex surrogates are not $\mathcal{H}_{\textrm{lin}}$-calibrated in the adversarial setting for binary classification. \citet{Awasthi_Calibrn_Consist_adv_sl, Awasthi_Finer_calibrn_analysis} extended calibration results in an adversarial setting to other function classes such as generalized linear models and single-layer ReLU neural networks. 

This paper studies the binary classifiers with reject option robust to adversarial perturbations. In high-risk environments (such as medical diagnosis, finance, etc.), the ability to abstain from prediction and incur small costs for it is beneficial as compared to the cost of misclassification. Classifiers with ability to abstain are called \enquote{reject option} classifiers \citep{Chow_Optimum_Rec_err_Rej_Trd,Cortes_Learning_with_Rejn, Chenri_Calibrn_Multi_Reject}. While many studies address adversarial robustness in standard classification setting, no attention has been given to the adversarial robustness of reject option classifiers.
In \citet{ATRO,Chen_Jha_Stratified_Adv_Rob_Rejection}, authors propose an adversarial robust approach for reject option classification and validate it empirically. 
However, there is no attempt to analyze calibration for this domain. 

We analyze calibrated surrogates in a scenario where the inputs are adversarially perturbed, and the classification has an embedded reject option. To our knowledge, this is the foundational work for this domain from the standpoint of calibration analysis.  
\subsubsection*{Key Contributions}
\begin{itemize}
    \item We completely characterize surrogates, which are calibrated in the "adversarial-reject option setting" for linear classifiers. We prove that convex loss functions can not be calibrated surrogates in such scenarios. We also show that surrogate losses with quasi-concave conditional risk are not calibrated.
    \item We propose that Shifted Double Sigmoid Loss (DSL) and Shifted Double Ramp Loss (DRL) are potential candidates for calibrated surrogates and empirically show that they satisfy calibration conditions.
    \item We describe the adversarial training procedure using proposed loss functions. We experimentally validate our findings on a synthetic dataset that Shifted DSL and Shifted DRL exhibit robustness against adversarial perturbations. 
\end{itemize}
\section{Related Work} \label{Related Work}
\subsection{Calibrated Surrogates towards Robust Adversarial Classification}
Adversarial robust loss for binary classification is $\max _{\tilde{\mathbf{x}} \in \mathcal{U}(\mathbf{x})} \ell(f(\tilde{\mathbf{x}}),y)$ where $\ell$ is a loss function and $\mathcal{U}(\mathbf{x})$ is an uncertainty set around $\mathbf{x}$. For $\mathcal{U}(\mathbf{x}) = B_{2}(\mathbf{x},\gamma)$, \citet{Bao_CSL_Adv_Rob} show that convex surrogates are negatively calibrated to the adversarial robust loss for linear classifiers ($\mathcal{H}_{\textrm{lin}}$) and gave positive calibration results by introducing the quasi-concavity assumption on the conditional risk. \citet{Awasthi_Calibrn_Consist_adv_sl, Awasthi_Finer_calibrn_analysis} extended calibration results for generalized linear models and single-layer ReLU neural network.
\citet{Meunier_Consistency_Adv_Classfn} show that shifted odd losses are calibrated to the adversarial robust loss.
\subsection{Robustness towards Adversarial Attacks}
Adversarial training, a popular adversarial defense approach, involves adding adversarial examples into the training set. 
Fast gradient sign method \citep{Goodfellow_Explain_Adv_attk} for $l_{\infty}$-norm perturbations, projected gradient descent \citep{Madry_DL_Resist_Adv_Attk} are some popular algorithms based on adversarial training.
The approach in \citet{ML_LOO_Adv_exmp_Feature_Attrn} is based on detecting adversarial examples using distance-based approaches or by feature attrition. To fool the gradient-based adversarial attack methods, \citet{Deep_k_NN_Robust} add discrete or non-differentiable components into the model. \citet{Certified_rob_Lipschtiz_NN_Boolean} investigates certified $l_{\infty}$ robustness from the lens of representing Boolean functions. Randomized smoothing \citep{Cohen_certified_adv_rob_smoothing} is a technique to convert any classifier that classifies well under Gaussian noise into a new classifier
that is certifiably robust to adversarial perturbations.

\subsection{Reject Option Classification}
\citet{Chow_Optimum_Rec_err_Rej_Trd} was the seminal work to deal with this approach to classification problems. Generalized hinge loss \citep{Bartlett_Hinge_Loss_Reject}, double ramp loss \citep{ManwaniDoubleRL, Shah_Manwani_Rob_Rej_Lin_Prog}, max-hinge loss and plus-hinge loss \citep{Cortes_Learning_with_Rejn} etc. are some of the consistent loss functions for learning with rejection in binary setting. Algorithms using these loss functions are support vector machine (SVM) variants for reject option classifiers.
\citet{DBLP:conf/aaai/ShahM20,DBLP:conf/uai/KalraSM21} proposed a new consistent loss function called double sigmoid loss function for binary reject option classifier. \citet{Ramaswamy_Consistent_Multi_Reject, Chenri_Calibrn_Multi_Reject, NEURIPS2022_03a90e1b} deal with calibration and consistency in the multiclass reject option classification. 
\section{Preliminaries} \label{Calibration Section}
\subsection{Notations}
For a vector $\x \in \mathbb{R}^{d}$, let $\| \x\|_{p}$ denote the $l_{p}$-norm. $B_{p}(\x,r) \stackrel{\text { def }}{=}\{\mathbf{v} \in$ $\left.\mathbb{R}^{d} \mid\|\mathbf{v} - \x \|_{p} \leq r\right\}$ be the $d$-dimensional closed $l_{p}$-ball centered at $\x$ with radius $r$. The set $\{1, \ldots, n\}$ is denoted by $[n]$. The indicator function corresponding to an event $A$ is denoted by $\mathbbm{1}_{\{A\}}$.

\subsection{Binary Classification Problem}
\textcolor{red}{} Let $\mathcal{X} \subseteq \mathbb{R}^d$ be the instance space and $\mathcal{Y}=\{1,-1\}$ be the label space. Let $\mathcal{P}$ be a fixed but unknown probability distribution over $\mathcal{X} \times \mathcal{Y}$ from which i.i.d.
samples of $(\mathbf{x},y)$ are drawn. The objective of the classification problem is to learn a function $f:\mathcal{X}\rightarrow \mathbb{R}$. For any example, the class label is predicted as $\hat{y}=\mathrm{sign}(f(\mathbf{x}))$. Loss $\ell_{01}$ captures the difference between the predicted label and the true label, where $\ell_{01}(yf( \mathbf{x}))=\mathbbm{1}_{y f(\mathbf{x}) \leq 0}$. 
The objective of the learning algorithm is to find a classifier $f^{*}$ in the function class $\mathcal{H}$ which minimizes the risk function $\mathcal{R}_{\ell_{01}}(f)=\mathbb{E}_{(\mathbf{x}, y) \sim \mathcal{P}}\left[\ell_{01}(yf(\mathbf{x}))\right]$. The risk $\mathcal{R}_{\ell_{01}}(f)$ is minimized by Bayes classifier $f^*(\mathbf{x})=\eta-\frac{1}{2}$, where $\eta=P(Y=1|\mathbf{x})$. In practice, we use surrogates of $\ell_{01}$ loss which are easy to optimize. The true risk of a classifier $f$ for a surrogate loss $\ell(yf(\mathbf{x}))$ is $
\mathcal{R}_{\ell}(f)=\underset{(\mathbf{x}, y) \sim \mathcal{P}}{\mathbb{E}}[\ell(yf(\mathbf{x}))]
$. The Bayes $(\ell, \mathcal{H})$-risk is defined by $\mathcal{R}_{\ell, \mathcal{H}}^{*}=$ $\inf _{f \in \mathcal{H}} \mathcal{R}_{\ell , \mathcal{H}}(f)$.

\subsection{Consistency of Surrogate Loss Functions}
\begin{definition} 
\textbf{$\mathcal{H}$-Consistency} : For a given hypothesis set $\mathcal{H}$ and a target loss function $\ell_{1}$, a surrogate $\ell_{2}$ is said to be $\mathcal{H}$-consistent with respect to $\ell_{1}$, if the following holds:
\begin{equation} \label{consistency}
\mathcal{R}_{\ell_{2}}\left(f_{n}\right)-\mathcal{R}_{\ell_{2}, \mathcal{H}}^{*} \stackrel{n \rightarrow+\infty}{\longrightarrow} 0 \Longrightarrow \mathcal{R}_{\ell_{1}}\left(f_{n}\right)-\mathcal{R}_{\ell_{1}, \mathcal{H}}^{*} \stackrel{n \rightarrow+\infty}{\longrightarrow} 0
\end{equation}
for all probability distributions and sequences of $\left\{f_{n}\right\}_{n \in \mathbb{N}} \subset \mathcal{H}$.
\end{definition}

Using conditional expectation property, $\mathcal{R}_{\ell}(f)$ can be written as $\mathcal{R}_{\ell} (f)=\mathbb{E}_{X}[\mathcal{C}_{\ell,\mathcal{H}}\left(f(\mathbf{x}), \eta\right)]$, where $\eta(\x) = P(y=1|\mathbf{x})$\footnote{For the rest of the paper, we adopt the notation as $\eta$ for $\eta(\x)$.} and
\begin{align}
\label{conditional_risk}
 \mathcal{C}_{\ell,\mathcal{H}}(f(\mathbf{x}), \eta)=\mathbb{E}_{y|\mathbf{x}}[\ell(yf(\mathbf{x}))~|~\mathbf{x}\;]=\eta~\ell(f(\mathbf{x}))+(1-\eta)~\ell(-f( \mathbf{x})).
\end{align} 

The minimal conditional risk $\mathcal{C}_{\ell,\mathcal{H}}^{*} (\mathbf{x},\eta)$ \citep{Steinwart_Compare_Diff_Losses} and pseudo-minimal 
conditional risk $\mathcal{C}_{\ell,\mathcal{H}}^{*}(\eta)$ \citep{Bao_CSL_Adv_Rob} are defined as 
\begin{equation}    \mathcal{C}_{\ell,\mathcal{H}}^{*} (\mathbf{x},\eta) = \inf_{f \in \mathcal{H}}~\mathcal{C}_{\ell , \mathcal{H}} (f(\mathbf{x}),\eta) \quad \textrm{and} \quad\mathcal{C}_{\ell,\mathcal{H}}^{*}(\eta) = \inf_{f \in \mathcal{H} , \mathbf{x} \in \mathcal{X}} \mathcal{C}_{\ell , \mathcal{H}} (f(\mathbf{x}),\eta)
\end{equation}
respectively. The corresponding excess-conditional risk is defined as : 
\begin{equation} \label{excess_conditional_risk}
    \Delta \mathcal{C}_{\ell , \mathcal{H}}(f(\mathbf{x}), \eta) = \mathcal{C}_{\ell, \mathcal{H}}(f(\mathbf{x}), \eta) - \mathcal{C}^{*}_{\ell, \mathcal{H}}(\mathbf{x},\eta) 
\end{equation}

\subsection{Calibration Theory}
\begin{definition}
    \textbf{Uniform $\mathcal{H}$-Calibration} [Definition 2.15 in \citep{Steinwart_Compare_Diff_Losses}] : For a given hypothesis set $\mathcal{H}$ and a target loss function $\ell_{1}$, a surrogate $\ell_{2}$ is said to be uniformly $\mathcal{H}$-calibrated with respect to $\ell_{1}$ if, for any $\epsilon>0$, there exists $\delta>0$ such that for all $\eta \in [0,1], f \in \mathcal{H}, \mathbf{x} \in \mathcal{X}$, we have
$
\mathcal{C}_{\ell_{2} , \mathcal{H}}(f( \mathbf{x}), \eta) - \mathcal{C}_{\ell_{2}, \mathcal{H}}^{*}(\mathbf{x}, \eta) < \delta \Longrightarrow \mathcal{C}_{\ell_{1} , \mathcal{H}}(f(\mathbf{x}), \eta)- \mathcal{C}_{\ell_{1}, \mathcal{H}}^{*}(\mathbf{x}, \eta) < \epsilon
$.
\end{definition}

\begin{definition}
\textbf{Uniform Pseudo- $\mathcal{H}$- Calibration} \citep{Bao_CSL_Adv_Rob} : For a given hypothesis set $\mathcal{H}$ and a target loss function $\ell_{1}$,  a surrogate loss function $\ell_{2}$ is uniformly pseudo- $\mathcal{H}$-calibrated with respect to a $\ell_{1} $ if, for any $\epsilon>0$, there exists $\delta>0$ such that for all $\eta \in[0,1]$ and $f \in \mathcal{H}, \mathbf{x} \in \mathcal{X}$, we have
$
\mathcal{C}_{\ell_{2} , \mathcal{H}}(f(\mathbf{x}), \eta)-\mathcal{C}_{\ell_{2}, \mathcal{H}}^{*}(\eta) < \delta \Longrightarrow \mathcal{C}_{\ell_{1} , \mathcal{H}}(f(\mathbf{x}), \eta) - \mathcal{C}_{\ell_{1}, \mathcal{H}}^{*}(\eta) < \epsilon
$.
\end{definition}

\begin{definition} \label{calibration_function}
   \textbf{Uniform Calibration function} \citep{Steinwart_Compare_Diff_Losses} : Given a hypothesis set $\mathcal{H}$, we define the uniform calibration function $\delta$ and uniform pseudo-calibration function $\hat{\delta}$ for a pair of losses $\left(\ell_{1}, \ell_{2}\right)$ as follows: for any $\epsilon>0$
$$
\begin{aligned}
& \delta(\epsilon)=\inf _{\eta \in[0,1]} \inf _{f \in \mathcal{H}, \mathbf{x} \in \mathcal{X}}\left\{\mathcal{C}_{\ell_{2},\mathcal{H}}(f(\mathbf{x}), \eta)-\mathcal{C}_{\ell_{2}, \mathcal{H}}^{*}(\mathbf{x}, \eta) \mid \mathcal{C}_{\ell_{1}, \mathcal{H}}(f(\mathbf{x}), \eta)-\mathcal{C}_{\ell_{1}, \mathcal{H}}^{*}(\mathbf{x}, \eta) \geq \epsilon\right\} \\
& \hat{\delta}(\epsilon)=\inf _{\eta \in[0,1]} \inf _{f \in \mathcal{H}, \mathbf{x} \in \mathcal{X}}\left\{\mathcal{C}_{\ell_{2} , \mathcal{H}}(f(\mathbf{x}), \eta)-\mathcal{C}_{\ell_{2}, \mathcal{H}}^{*}(\eta) \mid \mathcal{C}_{\ell_{1}, \mathcal{H}}(f(\mathbf{x}), \eta)-\mathcal{C}_{\ell_{1}, \mathcal{H}}^{*}(\eta) \geq \epsilon\right\} .
\end{aligned}
$$
\end{definition}

\begin{proposition} \label{calibration_prop}
[Lemma 2.16 in \citep{Steinwart_Compare_Diff_Losses}] Given a hypothesis set $\mathcal{H}$, loss $\ell_{2}$ is uniformly $\mathcal{H}$-calibrated (or uniformly pseudo-$\mathcal{H}$-calibrated) with respect to $\ell_1$ if and only if its calibration function $\delta$ satisfies $\delta(\epsilon)>0$ (resp. its uniform pseudo-calibration function $\hat{\delta}$ satisfies $\hat{\delta}(\epsilon)>0)$ for all $\epsilon>0$.
\end{proposition} 

For the standard binary classification problem, when $\mathcal{H} = \mathcal{H}_{all}$, \citet{Bartlett_Jordan_Conv_Clssfn_RB} showed that calibration is both necessary and sufficient for consistency. However, when we restrict ourselves to a hypothesis set ($\mathcal{H} \neq \mathcal{H}_{all}$), then calibration is a necessary but not sufficient condition for consistency. \citet{Steinwart_Compare_Diff_Losses} showed that if the loss functions satisfy an additional criteria of $\mathcal{P}$-minimizability, then point-wise minimization of conditional risk (calibration) does yield minimization of true risk (consistency).

\subsection{Calibration with Adversarial Robustness}
In the scenario when inputs are adversarially perturbed, designing surrogates that are calibrated and consistent becomes difficult. \citet{Bao_CSL_Adv_Rob} showed that convex losses are not calibrated when the hypothesis set consists of linear classifiers. The reason is that the convexity assumption on the surrogate results in the minimizer of the conditional risk being close to the origin (i.e., the non-robust region) for the case when the posterior probability for both classes is close to half ($\eta \approx 0.5)$. The robust loss in that case was defined as $\phi_{\gamma}(f(\mathbf{x}))= \mathbbm{1}_{\{y f(\mathbf{x}) \leq \gamma\}}$. Negative Calibration result is stated as follows : 

\begin{theorem} \label{Bao_Neg_calibrn_cvx_losses}\citep{Bao_CSL_Adv_Rob}
     For any margin-based surrogate loss $\ell: \mathbb{R} \rightarrow \mathbb{R}_{\geq 0}$, if $\ell$ is convex, then $\ell$ is not pseudo-calibrated $w r t \left(\phi_{\gamma}, \mathcal{H}_{\text{lin}}\right)$.
\end{theorem}

\subsection{Reject Option Classifier} 
A confidence-based binary reject option classifier \citep{Cortes_Learning_with_Rejn} is comprised of a function $f:\mathcal{X}\rightarrow \mathbb{R}$ and a confidence parameter $\rho \in \mathbb{R}_+$. The confidence-based reject option classifier is defined as
$h(f(\mathbf{x}),\rho)=
        1\;\mathbbm{1}_{\{f(\mathbf{x}) > \rho\}}\;+\;\perp\;\mathbbm{1}_{\{\vert f(\mathbf{x})\vert \leq \rho\}}\;+\;
        -1\;\mathbbm{1}_{\{f(\mathbf{x})<-\rho\}}$.
The Reject option loss $\ell_{d}$ for the \textbf{Confidence-based Rejection model} is defined as : 
\begin{equation} \label{0-d-1 defn}
     \ell_{d} (yf(\mathbf{x}),\rho) = \mathbbm{1}_{\{y f \left(\mathbf{x}\right )  \leq \textbf{-} \rho\}} + d\; \mathbbm{1}_{\{ | f (\mathbf{x}  )|   \leq \rho\}}
 \end{equation}
 where $d\in (0,0.5)$ is the cost of rejection.
 The generalized Bayes classifier \citep{Chow_Optimum_Rec_err_Rej_Trd} is defined as $
    f^{*}_{d}(\mathbf{x})=
        \mathbbm{1}_{\{\eta(\textbf{x}) > 1-d\}}\;+\;
    \perp\;\mathbbm{1}_{\{d \leq \eta(\textbf{x}) \leq 1-d\}}\;
        -\;\mathbbm{1}_{\{\eta(\textbf{x}) < d\}}$.

\section{Proposed Work: Calibration in the Adversarial Robust Reject Option Setting} \label{Proposed Work}

In our analysis, we assume $\mathcal{X}=B_{2}(\mathbf{0},1)$ ($l_2$ unit ball centered at origin). To start, we rewrite loss $\ell_{d}$ (see eq.~(\ref{0-d-1 defn})) as a convex combination of two indicator functions as follows. 
 \begin{equation} \label{new 0-d-1 defn}
    \ell_{d}(yf(\mathbf{x}),\rho) = (1-d) \; \mathbbm{1}_{ \{ y f\left(\mathbf{x} \right ) < \textbf{-} \rho\} } + d \; \mathbbm{1}_{ \{ y f\left(\mathbf{x} \right ) \leq \rho\} }
\end{equation}
An adversarial robust loss corresponding to $\ell_d$ is $\sup _{\mathbf{x}^{\prime}:\left\|\mathbf{x}-\mathbf{x}^{\prime}\right\|_2 \leq \gamma}  \ell_{d}(yf(\mathbf{x}^{\prime}),\rho)$. However, analysis of this loss is not easy. So, we define a new adversarial robust loss for confidence based reject option classifier. 

\begin{definition} \label{adv_rej_rob_loss}
    \textbf{Adversarial Robust Reject Option Loss: }Given $d$, $f$ and $\rho$, the adversarial reject option loss $\ell_d^{\gamma}$ for $(\mathbf{x},y)$ is defined as 
    \begin{align}
    \label{simplified-loss}
       \ell_d^{\gamma}(yf(\mathbf{x}),\rho)= (1-d) \; 
     \sup_{\mathbf{x}^{\prime}: \left\|\mathbf{x}-\mathbf{x}^{\prime}\right\|_2 \leq \gamma} \{  \mathbbm{1}_{ \{ y f\left(\mathbf{x}^{\prime} \right ) < -\rho\} } \} + \; d\; \sup _{\mathbf{x}^{\prime}:\left\|\mathbf{x}-\mathbf{x}^{\prime}\right\|_2 \leq \gamma} \{  \mathbbm{1}_{ \{ y f\left(\mathbf{x}^{\prime} \right ) \leq \rho\} } \}
    \end{align}
\end{definition}
Note that, in $\ell_d^{\gamma}(yf(\mathbf{x}),\rho)$, we consider $l_2$ norm perturbations. It is easy to see that $\ell_d^{\gamma}(yf(\mathbf{x}),\rho)\geq \sup _{\mathbf{x}^{\prime}:\left\|\mathbf{x}-\mathbf{x}^{\prime}\right\| \leq \gamma}  \ell_{d}(yf(\mathbf{x}^{\prime}),\rho)$.
We use $\ell_d^{\gamma}$ as target loss function. In this paper, we want to characterize the conditions under which surrogate loss functions are calibrated to the target loss function $\ell_d^{\gamma}$. The class of linear classifiers is defined as $\mathcal{H}_{\text{lin}} = \{ \mathbf{x} \rightarrow \mathbf{w} \cdot \mathbf{x} \;\mid \; \| \mathbf{w} \| = 1\}$. Now onwards, we consider $\mathcal{H} = \mathcal{H}_{\textrm{lin}}$. For linear classifiers, loss $\ell_{d}^{\gamma}$ becomes $\gamma$-right shift of $\ell_{d}$ loss which is shown in the next proposition.

\begin{proposition}
\label{prop1}
    The Adversarial Robust Reject Option Loss $\ell_{d}^{\gamma}$ for the class of linear classifiers is $\gamma$-right shift of $\ell_{d}$ loss as follows.
    \begin{equation} \label{adv_rob_rej_loss_linear_case}
           \ell^{\gamma}_{d}(yf(\mathbf{x}),\rho) = (1-d) \; \mathbbm{1}_{ \{  y f\left(\mathbf{x} \right ) < -\rho + \gamma \; \}} + \; d \; \mathbbm{1}_{ \{ \;  y f\left(\mathbf{x} \right ) \leq \;\rho +\gamma \; \}}
    \end{equation}
\end{proposition}

\subsection{Analysis of the Calibration Function} 
\begin{figure}[htp] 
\begin{tabular}{cc}
    \centering
\includegraphics[scale=0.4]{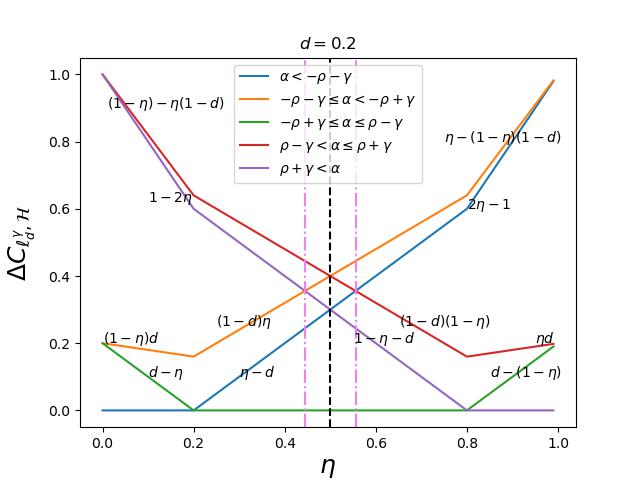} & \includegraphics[scale = 0.4]{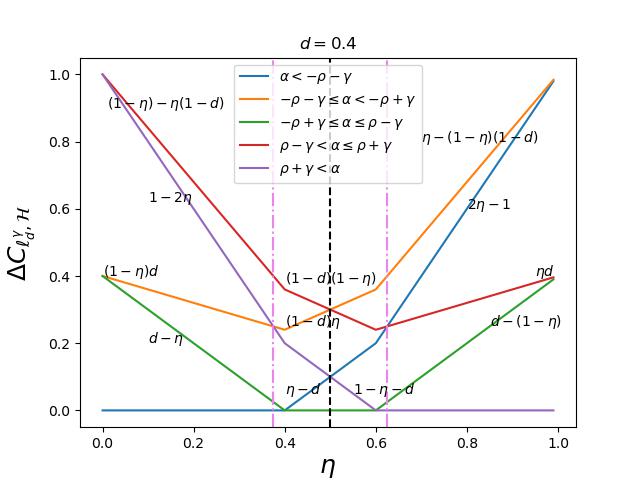} \\
    (a)&(b)
\end{tabular}
    \caption{Graph of excess target risk vs $\eta$ for two different $d$ values.}
    \label{excess_target_risk vs eta-plot}
\end{figure}

Here, we derive the calibration function for  
any margin based surrogate $\ell$ of loss $\ell^{\gamma}_{d}$. The inner risk $\mathcal{C}_{\ell_{d}^{\gamma} , \mathcal{H}}$ for $\ell^{\gamma}_{d}$ can be written as 
\begin{equation}    \mathcal{C}_{\ell_{d}^{\gamma} , \mathcal{H}} (f(\mathbf{x}),\eta) = \eta\;\ell_{d}^{\gamma}(f(\mathbf{x}),\rho) + (1-\eta)\;\ell_{d}^{\gamma}(-f(\mathbf{x}),\rho)
\end{equation}
Let $\alpha = f(\mathbf{x}) = \mathbf{w \cdot x}$. Piece-wise definition of the inner-risk for target loss  $\ell_{d}^{\gamma}$ is as follows.
\begin{align}
\nonumber \mathcal{C}_{\ell^{\gamma}_{d} , \mathcal{H}}(\alpha, \eta)&=
        \eta \;\mathbbm{1}_{\{\alpha < -\rho -\gamma\}}\;+\; 
        (\eta + (1-\eta)d)\; \mathbbm{1}_{\{ -\rho -\gamma \leq \alpha < -\rho + \gamma\}}\; +\;
        d\; \mathbbm{1}_{\{  -\rho +\gamma \leq \alpha \leq \rho - \gamma\}}\\
        &\;\;+\;
        (\eta d + (1-\eta)) \;\mathbbm{I}_{\{ \rho -\gamma < \alpha \leq  \rho  + \gamma\}}\;+\;
        (1-\eta) \;\mathbbm{I}_{\{  \rho + \gamma < \alpha\}}
\end{align}
We now find the excess inner risk $\Delta \mathcal{C}_{\ell^{\gamma}_{d} , \mathcal{H}}(\alpha, \eta)=\mathcal{C}_{\ell^{\gamma}_{d} , \mathcal{H}}(\alpha, \eta)-\mathcal{C}_{\ell^{\gamma}_{d} , \mathcal{H}}^{*}(\alpha,\eta)$ of the target loss $\ell_d^{\gamma}$. The following lemma gives a case by case expression for the same. 
\begin{lemma}
\label{lemma:excess-risk-target}
The excess-inner risk  for target loss $\ell^{\gamma}_{d}$ is given by
\begin{align*}
&\Delta \mathcal{C}_{\ell^{\gamma}_{d} , \mathcal{H}}(\alpha, \eta)=\mathcal{C}_{\ell^{\gamma}_{d} , \mathcal{H}}(\alpha, \eta)-\mathcal{C}_{\ell^{\gamma}_{d} , \mathcal{H}}^{*}(\alpha,\eta)=\\
&    \begin{cases}
        (\eta - d)\; \mathbbm{1}_{\min \{\eta,1-\eta \} - d \geq 0} + | 2\eta -1 |\;\mathbbm{1}_{2\eta -1 >0}\;\mathbbm{1}_{\min \{\eta,1-\eta\} - d < 0} & \textrm{if } \alpha < -\rho -\gamma \\ \hline
        (1- d)\;\eta \; \mathbbm{1}_{\min \{\eta,1-\eta \} - d \geq 0} + \; 
        ~\\
        \{ (\eta - (1-\eta)(1-d)) \;\mathbbm{1}_{2\eta -1 > 0} + (1-\eta)d\; \mathbbm{1}_{2\eta - 1< 0} \} \mathbbm{1}_{\min \{\eta,1-\eta\} - d < 0}   & \textrm{if }  -\rho -\gamma \leq \alpha  < -\rho + \gamma \\
        \hline
        \{(d - (1-\eta))\;\mathbbm{1}_{2\eta -1 >0} + (d-\eta)\;\mathbbm{1}_{2\eta -1 <0}\} \mathbbm{1}_{\min \{\eta,1-\eta\} - d < 0}  & \textrm{if }  -\rho +\gamma \leq \alpha \leq \rho - \gamma \\ 
        \hline
        (1- d)\;(1-\eta) \; \mathbbm{1}_{\min \{\eta,1-\eta \} - d \geq 0} +  \;  
        ~\\
        \{ \eta\;d \;\mathbbm{1}_{2\eta -1 > 0} + ((1-\eta) - \eta \;(1-d))\; \mathbbm{1}_{2\eta - 1< 0} \} \mathbbm{1}_{\min \{\eta,1-\eta\} - d < 0}   & \textrm{if }  \rho -\gamma < \alpha \leq  \rho  + \gamma \\ 
        \hline
        (1-\eta - d)\; \mathbbm{1}_{\min \{\eta,1-\eta \} - d \geq 0} + | 2\eta -1 |\;\mathbbm{1}_{2\eta -1 < 0}\;\mathbbm{1}_{\min \{\eta,1-\eta\} - d < 0}  & \textrm{if }  \rho + \gamma < \alpha 
    \end{cases}
\end{align*}
\end{lemma}
Figure~\ref{excess_target_risk vs eta-plot} illustrates plots of $\Delta \mathcal{C}_{\ell_{d}^{\gamma}, \mathcal{H}}$ vs $\eta$ for $d=0.2$ and $d=0.4$.  
The vertical lines in violet correspond to $\eta_{\text{left}} = \frac{1-d}{2-d} \;$ and $\eta_{\text{right}} = \frac{1}{2-d}$. We see that $\eta_{\text{left}}$ is the intersection of the cases $\rho + \gamma < \alpha $ and $-\rho -\gamma \leq \alpha < -\rho + \gamma$. $\eta_{\text{right}}$ is the intersection of the cases $ \alpha < -\rho - \gamma $ and $\rho -\gamma < \alpha \leq \rho + \gamma$.
Region change precedes definition change when $d < \eta_{\text{left}}$. At $d = \eta_{\text{left}}$, i.e when $d = \frac{3-\sqrt{5}}{2}\; \approx 0.38$, they coincide and for $d > \eta_{\text{left}}$ , definition change occurs before region change.
The above graphs highlight the cases when $d < \eta_{\textrm{left}} (d=0.2)$ and $d > \eta_{\textrm{left}} (d=0.4)$. These were used to develop the idea for splitting the rejection case into further sub-cases.
We observe following properties of conditional inner risk and excess risk.

\paragraph{Symmetry Property of $\mathcal{C}_{\ell , \mathcal{H}}(\alpha,\eta)$ and $\Delta \mathcal{C}_{\ell , \mathcal{H}}(\alpha,\eta)$ for margin based loss functions: } 
    For any margin based loss function $\ell$, using eq.\eqref{conditional_risk}, we can see that $\mathcal{C}_{\ell , \mathcal{H}}(\alpha,\eta) = \eta \; \ell(\alpha) + (1-\eta) \;\ell(-\alpha) = \mathcal{C}_{\ell , \mathcal{H}}(-\alpha,1-\eta)$. More specifically, for $\eta=\frac{1}{2}$, we can see that $\mathcal{C}_{\ell , \mathcal{H}}(\alpha,\frac{1}{2}) = \mathcal{C}_{\ell , \mathcal{H}}(-\alpha,\frac{1}{2})$. Thus, we can conclude that $\mathcal{C}_{\ell , \mathcal{H}}(\alpha,\eta)$ is symmetric about $\eta = \frac{1}{2}$. 
    Similarly, one can show that $\Delta \mathcal{C}_{\ell , \mathcal{H}}(\alpha,\eta)=\Delta \mathcal{C}_{\ell , \mathcal{H}}(-\alpha,1-\eta)$. Hence, $\Delta \mathcal{C}_{\ell , \mathcal{H}}(\alpha,\frac{1}{2})=\Delta \mathcal{C}_{\ell , \mathcal{H}}(-\alpha,\frac{1}{2})$, making $\Delta \mathcal{C}_{\ell , \mathcal{H}}$ symmetric about $\eta=\frac{1}{2}$.
Using this, we now characterize calibration of a margin based surrogate $\ell$ to $\ell_{d}^{\gamma}$ for linear classifiers.

\begin{theorem}
\label{thm1}
Any margin-based surrogate $\ell$ is ($\ell_{d}^{\gamma} , \mathcal{H}$)-calibrated if and only if it satisfies the following : 
\begin{align} \label{main_calibrn_cond_1}
   &\inf_{\rho - \gamma < \alpha \leq \| \mathbf{x} \| } \mathcal{C}_{\ell, \mathcal{H}} (\alpha,\frac{1}{2}) > \inf_{0 \leq \alpha \leq \| \mathbf{x} \|} \mathcal{C}_{\ell, \mathcal{H}} (\alpha,\frac{1}{2}) \\
 \label{main_calibrn_cond_2}
&\inf_{ - \| \mathbf{x} \| \leq \alpha \leq \rho + \gamma} \mathcal{C}_{\ell, \mathcal{H}}(\alpha,\eta) > \inf_{-\| \mathbf{x} \| \alpha \leq \| \mathbf{x} \|} \mathcal{C}_{\ell, \mathcal{H}}(\alpha,\eta)\;\;\eta \in (\frac{1}{2}, 1]
\end{align}
\end{theorem}

\textbf{Minima Jump Requirement:} 
For $\eta = 0.5$, ($\ref{main_calibrn_cond_1}$) is the calibration condition which implies that minima should be close to origin, specifically in the interval $[0,\rho - \gamma]$. Calibration to hold for the case of $\eta > 0.5$, ($\ref{main_calibrn_cond_2}$) should be satisfied implying minima lies beyond $\rho + \gamma$. So, even for a small increase ($\xi$) in value of $\eta$, the minima needs to jump from a region closer to origin to the region lying on the rightmost end of the interval $[-1,1]$.

\subsection{Negative Calibration of Convex Surrogates to \texorpdfstring{$\ell_{d}^{\gamma}$}{}}
In the adversarial binary classification setting, convex surrogates show negative calibration result (Theorem $\ref{Bao_Neg_calibrn_cvx_losses}$). Reason being that for $\eta = 0.5$, minimizer for the conditional risk falls in the non-robust region $[ -\gamma , \gamma]$. However, $\eta = 0.5$ does not yield problems in the adversarial robust reject option case. It is evident from ($\ref{main_calibrn_cond_1}$) that minima of the conditional risk being closer to origin is in fact needed for calibration (minimizer must lie in $[0,\rho-\gamma]$). For the case $\eta > 0.5$, problems do arise as minimizer is needed at rightmost end of the interval $[-\| \mathbf{x} \| , \| \mathbf{x} \|\;]$. This gives a negative result for calibration as stated in  Theorem $\ref{thm:negative-result-convex-surrogates}$.
\begin{theorem}
\label{thm:negative-result-convex-surrogates}
    Let $\ell$ be a differentiable and convex margin based surrogate to $\ell_{d}^{\gamma}$. Then, $\ell$ is not ($\ell_{d}^{\gamma},\mathcal{H}$)-calibrated.
\end{theorem}

\subsection{Negative Calibration Result When Conditional Risk is Quasi-Concave}
Here, we see an important negative result that any surrogate loss of $\ell_d^\gamma$ whose conditional risk is quasi-concave, is not calibrated to $\ell_d^\gamma$.
    
\begin{theorem} \label{theorem_neg_calibrn}
No margin-based surrogate $\ell$ satisfying the property of quasi-concavity of the conditional risk $\mathcal{C}_{\ell,\mathcal{H}}(\alpha, \eta)$ in $\alpha , \; \forall \eta \in [0,1]$ is ($\ell_{d}^{\gamma},\mathcal{H}$)-calibrated.
\end{theorem}
This result is in contrast to \citet[Theorem 14]{Bao_CSL_Adv_Rob} in the context of adversarial binary classification setting (without reject option) which says that any surrogate which has quasi-concave conditional risk is calibrated. 
The above theorem highlights that results from calibration in adversarial setting doesn't hold straightaway upon adding the reject option. 
Hence, the extension to calibrated surrogates of $\ell_{d}^{\gamma}$ is highly non-trivial and challenging.

\textbf{NOTE:} Proof of all the propositions, lemmas and theorems given in this section are provided in the supplementary material.

\section{Possible Calibrated Surrogates for \texorpdfstring{$\ell_{d}^{\gamma}$}{}} \label{Possible Calibrated Surrogates}
In this section, we present two surrogate loss functions which exhibit the properties required to be $\mathcal{H}$-calibrated to $\ell_{d}^{\gamma}$, namely shifted double sigmoid loss and shifted double ramp loss.
\subsection{Shifted Double Sigmoid Loss}
Double sigmoid loss (DSL) \citep{DBLP:conf/aaai/ShahM20} was presented in the context of standard reject option classification in the non-adversarial setting. It is shown to be a calibrated surrogate to $\ell_d$ \citep{DBLP:conf/uai/KalraSM21}. Double sigmoid loss is defined as : 
\begin{equation} \label{ds_1}
    \ell_{\text{ds}}^{\mu}(yf(\mathbf{x}),\rho) = 2\;d\;\sigma(yf(\mathbf{x})-\rho) + 2\;(1-d)\;\sigma(yf(\mathbf{x})+\rho) 
\end{equation}
where $\sigma(a) = \frac{1}{1+e^{\mu a}}$ and $\mu > 0$. We conjecture that shifted DSL is a calibrated loss for $\ell_{d}^{\gamma}$.   \begin{definition}[{\textbf{Shifted Double Sigmoid Loss}}]
    Given the shift parameter $\beta>0$, we define the shifted double sigmoid loss as,
    \begin{equation} \label{shifted DSL defn}
    \ell_{\text{ds}}^{\mu , \beta}(y f(\mathbf{x}),\rho) = \ell_{\text{ds}}^{\mu}(yf(\mathbf{x}) - \beta,\rho) =2\;d\;\sigma(yf(\mathbf{x})-\beta-\rho) + 2\;(1-d)\;\sigma(yf(\mathbf{x})-\beta+\rho).
\end{equation}

\end{definition}
As $\ell_{d}^{\gamma}$ (eq.($\ref{adv_rob_rej_loss_linear_case}$)) is $\gamma$-right shifted version of $\ell_{d}$ (Proposition $\ref{prop1}$), for $\ell_{\text{ds}}^{\mu , \beta}$ (eq.($\ref{shifted DSL defn}$)) to be surrogate to $\ell_{d}^{\gamma}$, the shift ($\beta$), has to be at least $\gamma$ i.e $\beta \geq \gamma$. 
We can easily see that $\ell_{\text{ds}}^{\mu , \beta}$ is not a convex function of $yf(\mathbf{x})$.
Conditional risk associated with $\ell_{\text{ds}}^{\mu , \beta}$ can be written as 
\begin{equation} \label{cond_risk_ds}
    \mathcal{C}_{\ell_{\text{ds}}^{\mu , \beta} , \mathcal{H}} (f(\mathbf{x}),\eta) = \eta\;\ell_{\text{ds}}^{\mu , \beta} (f(\mathbf{x}),\rho) + (1-\eta)\;\ell_{\text{ds}}^{\mu , \beta} (-f(\mathbf{x}),\rho)  
\end{equation}

\subsection*{Analysis of conditional risk of $\ell_{\text{ds}}^{\mu , \beta}$ for varying $\beta$ values :}
\begin{figure}[htp]
\begin{tabular}{cc}
    \centering
\includegraphics[scale=0.4]{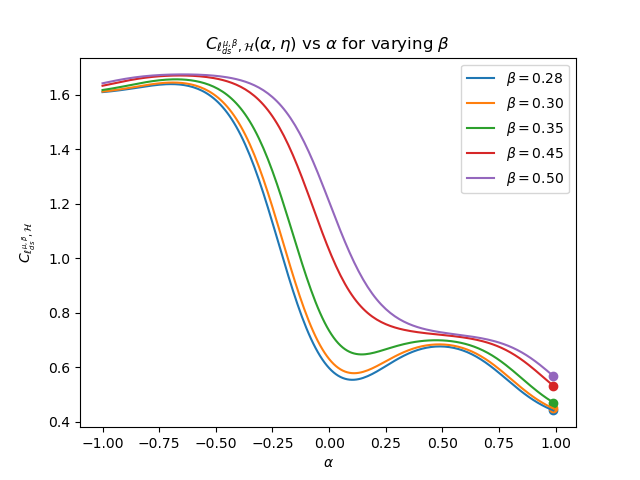} &\includegraphics[scale=0.4]{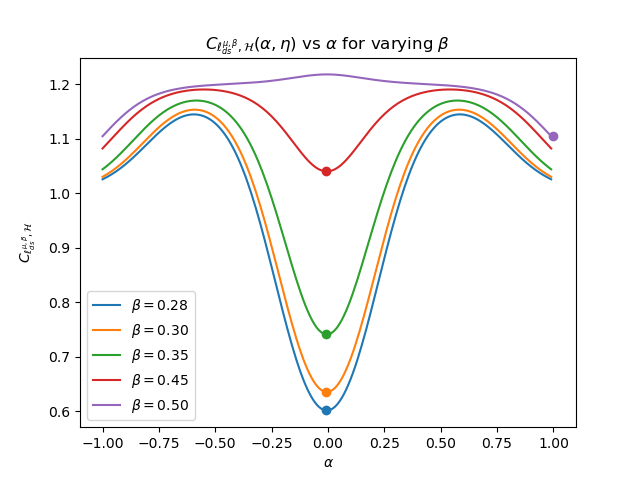}\\
(a) $\eta=0.6$ & (b) $\eta=0.5$
\end{tabular}
    \caption{(a) corresponds to case when $\eta =0.6$ and (b) corresponds to case when $\eta  =0.5$, for varying $\beta$ values with fixed $d=0.2$ and fixed $\mu=3.0$}
    \label{DS conditional risk vs alpha for varying beta}
\end{figure}

Here, we empirically demonstrate the effect of varying shift parameter ($\beta$) on the conditional risk to identify the cases when calibration holds. Figure~\ref{DS conditional risk vs alpha for varying beta} shows plots for varying $\beta$ for $\eta=0.6$ and $\eta = 0.5$ respectively.
We see that for fixed $\mu,d,\gamma$ - high $\beta$ values are needed to push the minima towards the right-most end (i.e close to $1$, see Figure~\ref{DS conditional risk vs alpha for varying beta}(a)) and low $\beta$ values keep minima closer to origin (Figure~\ref{DS conditional risk vs alpha for varying beta}(b)). High $\beta$ values are therefore favourable for ($\ref{main_calibrn_cond_2}$) and low $\beta$ values favour ($\ref{main_calibrn_cond_1}$). For calibration, both the conditions need to be satisfied. $\exists$ $\beta$ (depending on $\mu,\gamma,d$) for which these conditions are satisfied. 

 \subsection*{Analysis of conditional risk of $\ell_{\text{ds}}^{\mu , \beta}$ for varying $\eta$ values :} 
Here, we empirically demonstrate the effect of varying $\eta$ on the conditional risk to identify the cases when calibration holds.
Given below are the plot of ($\ref{cond_risk_ds}$) vs $\alpha$ for fixed $\beta = 0.45$ and fixed $d = 0.2$ with $\mu =3.0$ and $\mu = 2.65$ respectively.
\begin{figure}[htp] 
\begin{tabular}{cc}
    \centering
    \includegraphics[scale=0.4]{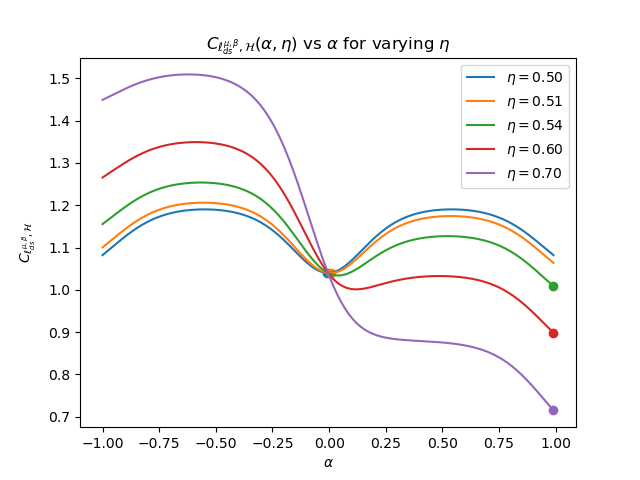} & \includegraphics[scale = 0.4]{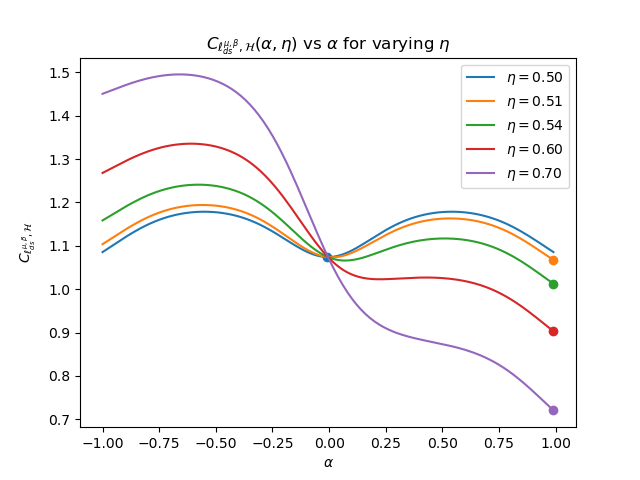} \\
    (a) $\mu = 3.0 $&(b) $\mu = 2.65$
\end{tabular}
    \caption{(a) corresponds to $\mu = 3.0$ and (b) corresponds to $\mu = 2.65$}
    \label{DS conditional risk vs alpha for varying eta}
\end{figure}
For the case when $\eta > 0.5$, we consider two offsets $\xi_{1} = 0.04$ and $\xi_{2} = 0.01$ and analyse calibration for $\eta + \xi_{1}$ and $\eta + \xi_{2}$. From Figure $\ref{DS conditional risk vs alpha for varying eta}$ (a), when $\mu = 3.0$, it is evident that minima for $\eta + \xi_{2}$ is located close to the origin, thereby violating ($\ref{main_calibrn_cond_1}$) whereas  minima for $\eta + \xi_{1}$ is located at the rightmost end. Upon reducing the eta value from $\eta + \xi_{1}$ to $\eta + \xi_{2}$, we need to reduce the value of $\mu$ so that calibration conditions are satisfied. This is evident from Figure $\ref{DS conditional risk vs alpha for varying eta}$ (b), where $\mu = 2.65$. We claim that $\exists$ a single $\mu$ value that satisfies both calibration conditions no matter how small the offset $\xi$ is taken. Empirically, it is seen that both calibration conditions ($\ref{main_calibrn_cond_1}$) and ($\ref{main_calibrn_cond_2}$) are satisfied. The minima lies in $[0,\rho-\gamma]$ for $\eta = 0.5$ and lies beyond $\rho+\gamma$ for $\eta > 0.5$.

\subsection{Shifted Double Ramp Loss} \label{DRL}
Double ramp loss (DRL) 
\citep{ManwaniDoubleRL} is proposed in the context of standard reject option classification without adversarial attacks. It is shown to be a calibrated surrogate to $\ell_d$ \citep{Shah_Manwani_Rob_Rej_Lin_Prog}. Let $[a]_{+} := \textrm{max}(0,a)$, then DRL is described as,
\begin{align}
\nonumber 
    \ell_{\text{dr}}^{\mu}&(yf(\mathbf{x}), \rho) = \frac{d}{\mu}   \left[  [\mu + \rho - yf(\mathbf{x})]_{+} - [-\mu^{2} + \rho - yf(\mathbf{x})]_{+} \right] + \frac{1-d}{\mu}  
    \left[  [\mu - \rho -yf(\mathbf{x}) ]_{+} - [-\mu^{2} -\rho - yf(\mathbf{x})]_{+}\right]
\end{align}


        



\begin{definition}[{Shifted Double Ramp Loss}]\label{shifted DRL defn} 
  Given the shift parameter $\beta (\geq 0)$, shifted double ramp loss is defined as $\ell_{\text{dr}}^{\mu , \beta} (yf(\mathbf{x}),\rho)=\ell_{\text{dr}}^{\mu} (yf(\mathbf{x})-\beta,\rho)$.
\end{definition} 

For shifted DRL $\ell_{\text{dr}}^{\mu , \beta}$ to be a surrogate for $\ell_{d}^{\gamma}$, we require that $\beta \geq \gamma$.  

Conditional risk associated with $\ell_{\text{dr}}^{\mu , \beta}$ is 
$\mathcal{C}_{\ell_{\text{dr}}^{\mu , \beta} , \mathcal{H}} (f(\mathbf{x}),\eta) = \eta\;\ell_{\text{dr}}^{\mu , \beta} (f(\mathbf{x}),\rho) + (1-\eta)\;\ell_{\text{dr}}^{\mu , \beta} (-f(\mathbf{x}),\rho)$.
\begin{figure}[htp] 
\label{}
\begin{tabular}{cc}
    \centering
    \includegraphics[scale=0.4]{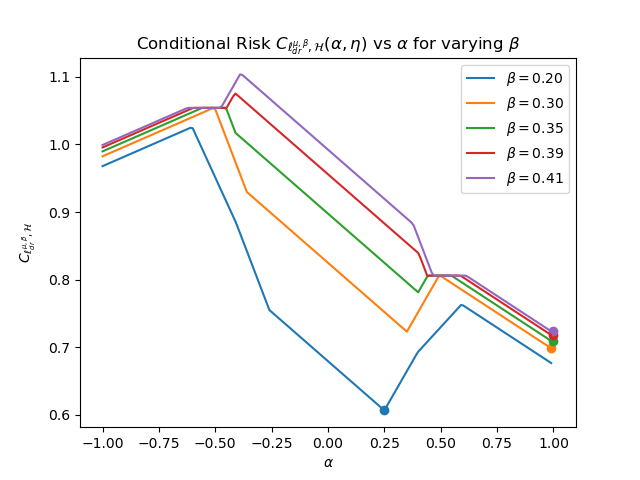} & \includegraphics[scale = 0.4]{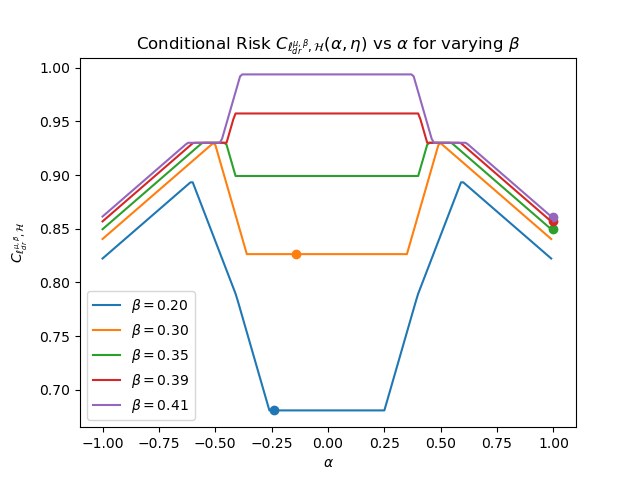} \\
    (a) $\eta = 0.6 $&(b) $\eta = 0.5$
\end{tabular}
    \caption{Conditional risk $\mathcal{C}_{\ell_{\text{dr}}^{\mu , \beta} , \mathcal{H}}$ versus shift parameter $\beta$ for $d , \mu, \gamma = \{0.2, 0.55, 0.2\}$ respectively. (a) For $\eta = 0.6$, high $\beta$ values push minima towards the rightmost end (beyond the $\rho+\gamma$ mark). (b) For $\eta = 0.5$, low $\beta$ (values that are closer to $\gamma$) are good to ensure that minima lies in $\left[ 0 , \rho - \gamma \right]$.}
    \label{DRL conditional risk vs alpha for varying beta}
\end{figure}
\subsection*{Analysis of Conditional risk of $\ell_{\text{dr}}^{\mu , \beta}$ for varying $\beta$ :}
Figure~\ref{DRL conditional risk vs alpha for varying beta} shows plots of $\mathcal{C}_{\ell_{\text{dr}}^{\mu , \beta} , \mathcal{H}}$ with varying $\beta$ values for fixed $d(=0.2),\mu(=0.55),\gamma(=0.2)$. 
We make following observations.
Low $\beta$ (values that are closer to $\gamma$) are good to ensure that minima lies in $\left[ 0 , \rho - \gamma \right]$ thereby, satisfying $\eqref{main_calibrn_cond_1}$ for $\eta = \frac{1}{2}$ (as seen in Figure $\ref{DRL conditional risk vs alpha for varying beta}$ (b)) but it violates  $\eqref{main_calibrn_cond_2}$ for $\eta > 0.5$ (as seen in Figure $\ref{DRL conditional risk vs alpha for varying beta}$ (a)). On the other hand, high $\beta$ values push minima towards the rightmost end (beyond the $\rho+\gamma$ mark), thereby favoring $\eqref{main_calibrn_cond_2}$ for $\eta > 0.5$ but violating $\eqref{main_calibrn_cond_1}$ for $\eta = 0.5$ case. Similar to shifted Double Sigmoid case, here too, $\exists$ $\beta$ value (depending on $\mu,d,\gamma$) which satisfies both calibration conditions.

\subsection*{Analysis of Conditional risk of $\ell_{\text{dr}}^{\mu , \beta}$ for varying $\eta$ }
Given below is the plot for varying $\eta$ for $\beta = 0.3$ and fixed $\mu,d,\gamma = 0.55,0.2,0.2$ respectively.
\begin{figure}[htp]
\begin{center} \label{DRL conditional risk vs alpha for varying eta}    \includegraphics[scale=0.4]{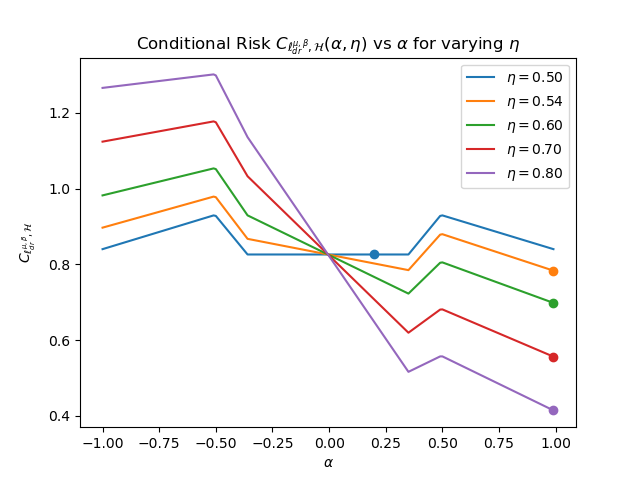}
\caption{conditional risk $\mathcal{C}_{\ell^{\mu,\beta}_{dr} , \mathcal{H}}$ vs $\alpha$ for varying $\eta$.}
\end{center}
\end{figure}
For shifted DRL also we observe "Minima-Jump". We see that, in the region around the origin, $\mathcal{C}_{\ell^{\mu,\beta}_{dr} , \mathcal{H}}$ remains constant. For every offset $\xi$ around $\eta$ that we consider, $\mu$ has to be adjusted while keeping $d,\gamma$ fixed to ensure calibration. As seen in Figure $\ref{DRL conditional risk vs alpha for varying eta}$, for $\xi = 0.04$, $\mu = 0.55$ with $\beta = 0.3$ yields calibration.

\subsection{Non Quasi-Concavity of \texorpdfstring{$\mathcal{C}_{\ell_{\text{ds}}^{\mu , \beta} , \mathcal{H}} (f(\mathbf{x}),\eta)$}{} and \texorpdfstring{$\mathcal{C}_{\ell_{\text{dr}}^{\mu , \beta} , \mathcal{H}} (f(\mathbf{x}),\eta)$}{}}
Theorem~\ref{theorem_neg_calibrn} states that if for a margin based surrogate of $\ell_d^\gamma$, the conditional risk is quasi-concave in $\alpha~~\forall \eta \in [0,1]$, then it is not ($\ell_{d}^{\gamma},\mathcal{H}$)-calibrated. It can be seen easily (refer Fig~\ref{DS conditional risk vs alpha for varying eta} and Fig~\ref{DRL conditional risk vs alpha for varying eta}) that $\mathcal{C}_{\ell_{\text{ds}}^{\mu , \beta} , \mathcal{H}} (f(\mathbf{x}),\eta)$ and $\mathcal{C}_{\ell_{\text{dr}}^{\mu , \beta} , \mathcal{H}} (f(\mathbf{x}),\eta)$ are not quasi-concave in $\alpha(=f(\x))~ \forall \eta$. This property makes $\ell_{\text{ds}}^{\mu , \beta}$ and $\ell_{\text{dr}}^{\mu , \beta}$ candidate surrogates which can be ($\ell_{d}^{\gamma},\mathcal{H}$)-calibrated.

\subsection{Adversarial Training using the Double Sigmoid Loss / Double Ramp Loss} \label{Adv Training using DSL/DRL}
In this section, we present a generic algorithm for adversarial training of linear reject option classifier using shifted DSL $\ell_{\textrm{ds}}^{\mu,\beta}$ and shifted DRL $\ell_{\textrm{dr}}^{\mu,\beta}$.
We explain here adversarial learning using shifted DSL $\ell_{\textrm{ds}}^{\mu,\beta}$. For shifted DRL, we adopt a similar approach.\\
{\textbf{\underline{Step 1}: } } Train a linear reject option classifier (parameters are  $\Theta=[\mathbf{w},\rho]^T$) on clean data $\mathcal{D}_{\textrm{clean}} = \{( \mathbf{x}_{i} , y_i)\}_{i=1}^N$ by minimizing empirical risk under $\ell_{\textrm{ds}}^{\mu}$. Bias term can be included in the vector $\mathbf{w}$ by appending 1 in the feature vector $\mathbf{x}$. 
Optimal parameters $\Theta^*=[\mathbf{w}^*,\rho^*]^T$ are obtained as,
$\Theta^*=\arg\min_{\Theta}\;\frac{1}{N}\sum_{i=1}^N \ell_{\textrm{ds}}^{\mu}(y_if(\mathbf{x}_i),\rho).$ using stochastic gradient descent. The optimal classifier is $f^*(\mathbf{x})=\mathbf{w}^*\cdot\mathbf{x}$.\\ {\textbf{\underline{Step 2}: } } Generate adversarial data over a subset of examples indexed by set $\mathcal{I}\subseteq [N]$. For any $\mathbf{x}_i,\;i\in\mathcal{I} \;\textrm{and}\; \gamma >0$, its adversarial corrupted version $\mathbf{x}_i^\gamma$ is generated as $\mathbf{x}^{\gamma}_i=\arg\max_{\mathbf{x}_{i}^{\prime}\in B_2(\mathbf{x}_i,\gamma)}\;\ell_{\textrm{ds}}^{\mu}(y_i\;f^*(\mathbf{x}_{i}^{\prime}),\rho^*).$ 
Projected Gradient Ascent is used to maximise $\eqref{ds_1}$,  by taking an ascent step in the gradient direction and projecting it onto $B_2(\mathbf{x}_i \gamma)$, in succession.

An adversarial dataset is created by adding the perturbed samples to clean data samples  $\mathcal{D}_{\textrm{adv}} = \{( \mathbf{x}_i^{\gamma},y_i)\}_{i=1}^{N} = \{(\x_{i}^{\gamma},y_i), \;i\in \mathcal{I}\} \cup \{( \x_{i},y_i) , i \in [N]\setminus \mathcal{I}  \}$. \\
{\textbf{\underline{Step 3}: } }Train a robust linear reject option classifier by minimizing the empirical risk on adversarial data $\mathcal{D}_{\textrm{adv}}$ using shifted DSL. The optimal parameters $\Theta^\gamma=[\mathbf{w}^\gamma,\rho^\gamma]^T$ after adversarial training are found as,
$\Theta^\gamma=\arg\min_{\Theta}\;\frac{1}{ N}\sum_{i=1}^{N}\ell_{\textrm{ds}}^{\beta,\mu}(y_i(\mathbf{w}\cdot \mathbf{x}_{i}^\gamma),\rho).$

\section{Experiments} \label{Experiments}

\subsection{Baselines and Dataset Description}
The linear Reject Option Classifiers (ROC) trained using DSL or DRL without shift ($\beta = 0$) are used as base models. Introducing shift ($\beta > 0$) in the corresponding DSL or DRL makes them robust to $l_{2}$-norm attacks. We report the performance of the non-robust $(\beta = 0)$ and three robust classifiers ($\beta = 0.1,0.15,0.25$) on it as shown in Table \ref{Synthetic Data table-DSL} and Table \ref{Synthetic Data table-DRL}. We do not choose ATRO \citep{ATRO} as a baseline as ATRO works on $l_{\infty}$ perturbations, whereas our work deals with $l_2$ perturbations. \\ 
We generate a linearly separable data ($\in \mathbb{R}^{2}$) with separation boundary as $\textbf{x} = 0$. All points should lie in $ B_{2}(\mathbf{0},1)$, the unit circle centred at origin. Take rejection width $= 0.5$ and 100 points per class in the reject region and 200 points per class in the non-reject region. Flip the labels of 5\% of the samples in the reject region for each class. This data will be used to train the classifiers. For testing, we generate data similarly, except for half the count of training. The perturbation radius of $l_{2}$-attack, $\gamma$ is referred as $\gamma_{\textrm{train}}$ and $\gamma_{\textrm{test}}$ for training time and test-time respectively ; $\gamma_{\textrm{train}} \in \{ 0.0, 0.1, 0.2\}$ and $\gamma_{\textrm{test}} \in \{ 0.0, 0.1, 0.2\}$. Every classifier trained using any value of $\gamma_{\textrm{train}}$ is tested with all three values of $\gamma_{\textrm{test}}$.

\begin{table*}[!ht]
    \begin{center}
    \scalebox{0.8}[0.8]{
    \begin{tabular}{|l|l|l|lll|lll|lll|}
 \hline
  &&     &   \multicolumn{3}{c|}{Attack $\gamma_{\textrm{test}}=0$}    &   \multicolumn{3}{c|}{ $\gamma_{\textrm{test}}=0.1$} &  \multicolumn{3}{c}{ $\gamma_{\textrm{test}}=0.2$} \\
 \multirow{1}{*}{$\gamma_{\textrm{train}}$} &  \multirow{1}{*}{$d$} &  \multirow{1}{*}{Training Loss } &  Err &   Acc     &    RR & Err & Acc & RR &  Err &  Acc    & RR \\
\hline
\hline
$0$& 0.2 & DSL ($\beta = 0$) &    0.338  &  0.458 & 0.53   &0.38  & 0.394  &0.514  & 0.484 & 0.306 &0.41  \\
& &DSL ($\beta = 0.1$)    &0.315   & 0.353  &0.612   & 0.342 & 0.308 &0.604 &0.409  & 0.24 & 0.544 \\
& &DSL ($\beta = 0.15$) &0.251   &0.15   & 0.828  &0.26  &0.135  & 0.825& 0.28 & 0.116 & 0.808 \\
& & DSL ($\beta = 0.25 $)  &\textcolor{blue}{\textbf{ 0.25}}   & 0.148  &0.835   & \textcolor{blue}{\textbf{0.258}} & 0.132 &0.836 &\textcolor{blue}{\textbf{0.275}}  &0.107  &  0.829\\
      \cline{2-12}
       &0.3 &DSL ($\beta = 0$) & 0.435 &  0.501 &0.315   & 0.552  &0.38  &0.202  &0.663  &  0.296 & 0.106 \\
       & &DSL ($\beta = 0.1$)& 0.43 & 0.5 & 0.35  &0.525   & 0.394 & 0.264 &0.638  & 0.304  &0.15 \\
       & & DSL ($\beta = 0.15$)&  0.378 & 0.297 & 0.615  &0.430   & 0.237 &0.57  &0.509  & 0.174 & 0.488\\
       & & DSL ($\beta = 0.25$) &\textcolor{blue}{\textbf{0.375}}  & 0.297 & 0.631   & \textcolor{blue}{\textbf{0.407}}  &0.253  &0.613  &\textcolor{blue}{\textbf{ 0.486}}  &0.182  &0.529 \\
       \cline{2-12}
       &0.4 &DSL ($\beta = 0$)&  0.476&0.507  & 0.176  & 0.623 & 0.345& 0.109 & 0.7 & 0.288 & 0.035\\
       &&DSL ($\beta = 0.1$)&  0.475&0.506 &  0.194  &0.616  &0.353 &0.121  &0.699  &   0.288 &0.039 \\
       && DSL ($\beta = 0.15$) &\textcolor{blue}{\textbf{ 0.436}} & 0.448&0.343  &0.578 &0.32  &0.272  & 0.651 &  0.261  &0.197 \\
      &&DSL ($\beta = 0.25$)&  0.457& 0.398&0.432 &\textcolor{blue}{\textbf{0.553}} &0.291  &  0.348& \textcolor{blue}{\textbf{0.634}} & 0.226   & 0.259\\
      \hline
      \hline

$0.1$ & 0.2 &DSL ($\beta=0$)  &  0.363 & 0.45 &  0.456   &  0.414 & 0.38  & 0.434 & 0.537  &0.273  &0.327 \\
& &DSL ($\beta = 0.1$) &   0.36 & 0.447 &  0.469  & 0.393  &0.393   &0.467  & 0.49 & 0.298 &0.383 \\
& &DSL ($\beta = 0.15$) &  \textcolor{blue}{\textbf{0.285}} & 0.336 & 0.711  & \textcolor{blue}{\textbf{ 0.302}} &  0.304 &0.712  &\textcolor{blue}{\textbf{ 0.348}}  & 0.255 &0.678 \\
& &DSL ($\beta = 0.25$) &   0.302  & 0.296  & 0.664  &0.323  &0.257  & 0.667 & 0.388 &0.189  &0.611  \\
\cline{2-12}
      &0.3 &DSL ($\beta = 0$)&   0.437 & 0.498 &  0.319  &0.56  &0.367  & 0.216  &0.681 &0.277  & 0.097  \\
      &&DSL ($\beta = 0.1$)&  0.425 & 0.5 &0.372  & 0.527&0.38  &0.288 &0.64 & 0.29 & 0.176 \\
       &&DSL ($\beta = 0.15$) &0.426   &0.501  &0.36   &   0.524&0.389  & 0.277  &  0.65  & 0.287 &0.151  \\
        &&DSL ($\beta = 0.25$)&  \textcolor{blue}{\textbf{ 0.423}} & 0.5 & 0.38    &  \textcolor{blue}{\textbf{0.509}} & 0.396  & 0.313 &\textcolor{blue}{\textbf{0.627}}  &0.291   & 0.202 \\
        \cline{2-12}
      &0.4 & DSL ($\beta = 0$)&  0.485  & 0.499 &   0.155  &0.624  &  0.352&0.098  &0.698 &0.291  &0.034  \\
 &&DSL ($\beta = 0.1$)&  0.479   & 0.501 & 0.187  & 0.615 & 0.356 & 0.115 & 0.697  &0.290  &0.04  \\
&&DSL ($\beta = 0.15$)&  0.478  &0.498  &0.227   & 0.611  &0.354  &0.136  &  0.698 & 0.284  &0.053   \\
 &&DSL ($\beta = 0.25$)& \textcolor{blue}{\textbf{0.464}}  &0.455  & 0.324  & \textcolor{blue}{\textbf{0.583}} &0.321  &0.244  &\textcolor{blue}{\textbf{0.667}}  &  0.253   & 0.161\\
      \hline
      \hline
$0.2$ & 0.2 &DSL ($\beta = 0$)  &    0.341  &  0.404 & 0.518 & 0.382 & 0.347 &0.502 & 0.48 & 0.264 &0.405  \\
&& DSL ($\beta = 0.1$) &    \textcolor{blue}{\textbf{ 0.2}}  &  0 & 1 &\textcolor{blue}{\textbf{ 0.2}}  &0  & 1&\textcolor{blue}{\textbf{ 0.2}}  &0  &1  \\
& & DSL ($\beta = 0.15$) &  0.218  &  0.048 &  0.942&  0.219  &0.047  &0.942 &0.219  &0.047  &0.942  \\
& &DSL ($\beta = 0.25$) &  0.218  &  0.048 & 0.942 & 0.218 & 0.048 &0.942 &0.218  & 0.048 &0.942  \\
\cline{2-12}
       &0.3 &DSL ($\beta = 0$)& 0.434  & 0.501 &  0.32  & 0.551  & 0.379 &0.217  &0.671  & 0.288 &0.098 \\
       &&DSL ($\beta = 0.1$)& 0.332 &0.203  &  0.833 &0.349   &0.166  &  0.825& 0.355 & 0.157 &0.816 \\
       &&DSL ($\beta = 0.15$) & 0.315 & 0.103 & 0.915  &0.323   &0.086  & 0.911 & 0.326 & 0.082 &0.906 \\
       &&DSL ($\beta = 0.25$)&  \textcolor{blue}{\textbf{ 0.3}} & 0 &  1 & \textcolor{blue}{\textbf{ 0.3}}  & 0 &1  & \textcolor{blue}{\textbf{ 0.3}} &0  &1 \\
       \cline{2-12}
      &0.4 &DSL ($\beta = 0$)& 0.477 &0.506 &0.168  &0.623 & 0.351 & 0.1 &  0.698&  0.29  &0.034 \\
 &&DSL ($\beta = 0.1$)& 0.437 & 0.5&0.628  & 0.465&0.435  & 0.633 &  0.479&   0.414 &0.612 \\
 &&DSL ($\beta = 0.15$)&  0.43&0.353 &0.698 & 0.453&0.3  & 0.702 &  0.462& 0.285   & 0.69\\
&&DSL ($\beta = 0.25$)& \textcolor{blue}{\textbf{ 0.409}} &0.09 &0.904  &\textcolor{blue}{\textbf{ 0.417}} & 0.08 & 0.906 & \textcolor{blue}{\textbf{0.418}} &0.08    &0.904 \\
\bottomrule
\end{tabular}}
 \caption{Results with linear reject option classifier with/without shift trained using Double Sigmoid Loss ($\mu = 2.65$).}
    \label{Synthetic Data table-DSL}
\end{center}
\end{table*}


\subsection{Observations}
Table~\ref{Synthetic Data table-DSL} and Table~\ref{Synthetic Data table-DRL} show results for shifted DSL and shifted DRL. For each value of $d,\mu,\gamma_{\textrm{train}},\beta$ ; we report the following metrics evaluated on test-data by averaging over 10 runs: (a) error (b) rejection rate  (c) accuracy on the predicted samples. 


\subsubsection{Effect of Increasing \texorpdfstring{$\gamma_{\textrm{test}}$}{} in the Test Time Attack}
For a fixed value of $\gamma_{\textrm{train}},\; d$ and $\beta$, as test-time attack $\gamma_{\textrm{test}}$ increases, the error increases. We observe this behavior with both shifted DSL and shifted DRL. This happens due to the following reason. By increasing the $\gamma_{\textrm{test}}$, the overlap between the two classes increases, increasing the linear classifier's error. However, we observe the following additional property in the case of shifted DRL. If we train with shifted DRL for a certain  $\gamma_{\textrm{train}}$, its error at test time does not increase much by increasing $\gamma_{\textrm{test}}$ as long as $\gamma_{\textrm{test}}\leq \gamma_{\textrm{train}}$. Shifted DRL has a flat region for $yf(\x) \in [-\rho + \mu+\beta,\rho - \mu^2+\beta]\cup [-\infty, -\rho-\mu^2+\beta]\cup[\rho+\mu+\beta,\infty)$. For smaller $\gamma_{\textrm{test}}$, pushing out an example out of these flat regions is hard. For sufficiently large $\gamma_{\textrm{test}}$, the loss can increase for such points in two ways. (a) Correctly classified data point in the region $[\rho+\mu+\beta,\infty)$ after $\gamma_{test}$ perturbation leaves the zero loss region and moves towards rejection region, thereby increasing loss value. (b) The data point is in the region $[-\rho + \mu+\beta,\rho - \mu^2+\beta]$ where the loss value is $d(1+\mu)$, and after perturbation moves towards the misclassification region and achieves higher loss values. 

 \subsubsection{Effect of Increasing \texorpdfstring{$d$}{}}
 For any robust classifier (with fixed $\beta, \gamma_{\textrm{train}}$), an increase in the cost of rejection $d$ leads to an increase in error and a reduction in the rejection rate. However, at high values of training $\gamma\;(=0.2)$, the overlap between the two classes becomes very large, and the classifier starts rejecting almost all samples. This is the common behavior of any reject option classifier.

 \subsubsection{Effect of Increasing \texorpdfstring{$\beta$}{}}
 Robust classifiers ($\beta = 0.1,0.15,0.25$) give less error than their non-robust ($\beta=0$) counterparts, which is expected from shifted DSL and shifted DRL. However, we observe this behavior when $\gamma_{\textrm{train}}$ is nonzero in the case of shifted DRL. For $\gamma_{\textrm{train}}=0$, changing $\beta$ does not make any change in the performance of shifted DRL for different values of $d$ and $\gamma_{\textrm{test}}$. Also, for fixed $d,\;\beta$ and for any $\gamma_{\textrm{test}} \leq \gamma_{\textrm{train}}$, the difference between the errors of non-robust classifier and the robust classifier is very small. This gap starts to widen when $\gamma_{\textrm{test}} > \gamma_{\textrm{train}}$. 

\begin{table*}[!ht]
    \begin{center}
    \scalebox{0.8}[0.8]{
    \begin{tabular}{|l|l|l|lll|lll|lll|}
 \hline
  &&     &   \multicolumn{3}{c|}{Attack $\gamma_{\textrm{test}}=0$}    &   \multicolumn{3}{c|}{Attack $\gamma_{\textrm{test}}=0.1$} &  \multicolumn{3}{c}{Attack $\gamma_{\textrm{test}}=0.2$} \\
 \multirow{1}{*}{ $\gamma_{\textrm{train}}$} &  \multirow{1}{*}{$d$} &  \multirow{1}{*}{Training Loss } &  Err &   Acc     &    RR & Err & Acc & RR &  Err &  Acc    & RR \\
\hline
$0$& 0.2 & DRL ($\beta = 0$) &  \textcolor{blue}{\textbf{0.451}}&0.494  &0.175       &0.538  &0.481  &0.114  & \textcolor{blue}{\textbf{ 0.545}} &0.414&0.101  \\
& &DRL ($\beta = 0.1$) & \textcolor{blue}{\textbf{0.451}}& 0.495 &0.173    & \textcolor{blue}{\textbf{ 0.537}}  &0.418  &0.113 &\textcolor{blue}{\textbf{0.545}} &0.414&0.101\\
& &DRL ($\beta = 0.15$) & \textcolor{blue}{\textbf{ 0.451}}& 0.495 &0.173    & \textcolor{blue}{\textbf{0.537}}  &0.418  &0.113 &\textcolor{blue}{\textbf{0.545}} &0.414&0.101\\
& & DRL ($\beta = 0.25 $) &  \textcolor{blue}{\textbf{ 0.451}}& 0.495 &0.173    & \textcolor{blue}{\textbf{0.537}}  &0.418  &0.113 &\textcolor{blue}{\textbf{0.545}} &0.414&0.101\\
      \cline{2-12}
       &0.3 &DRL ($\beta = 0$) &\textcolor{blue}{\textbf{ 0.475}}  &0.498  &0.129   & 0.561  &0.416  &0.082  &\textcolor{blue}{\textbf{0.559}}  &0.419  & 0.075\\
       & &DRL ($\beta = 0.1$)&\textcolor{blue}{\textbf{0.475}}  &0.498  &0.129   & \textcolor{blue}{\textbf{0.56}}  &0.416  &0.082   &\textcolor{blue}{\textbf{0.559}}  &0.419  & 0.075\\
       & & DRL ($\beta = 0.15$)&\textcolor{blue}{\textbf{0.475}}  &0.498  &0.129     & \textcolor{blue}{\textbf{0.56}}  &0.416  &0.082   &\textcolor{blue}{\textbf{0.559}}  &0.419  & 0.075\\
       & & DRL ($\beta = 0.25$) &\textcolor{blue}{\textbf{0.475}} &0.498  &0.129     & \textcolor{blue}{\textbf{0.56}}  &0.416  &0.082   &\textcolor{blue}{\textbf{0.559}}  &0.419  & 0.075\\
       \cline{2-12}
       &0.4 &DRL ($\beta = 0$)&\textcolor{blue}{\textbf{0.491}}  &0.497  &0.106   &0.574   &0.416  &0.065  & 0.565 &0.424  &0.058\\
       &&DRL ($\beta = 0.1$)&\textcolor{blue}{\textbf{ 0.491}}   &0.497  &0.106  &\textcolor{blue}{\textbf{0.573}}   &0.414  &0.064  & \textcolor{blue}{\textbf{0.564}} &0.424  & 0.058\\
       && DRL ($\beta = 0.15$)&\textcolor{blue}{\textbf{0.491}}   &0.497  &0.106   &\textcolor{blue}{\textbf{0.573}}   &0.414  &0.064  & \textcolor{blue}{\textbf{0.564}} &0.424  & 0.058\\
      &&DRL ($\beta = 0.25$)&\textcolor{blue}{\textbf{ 0.491}}  &0.497  &0.106   &\textcolor{blue}{\textbf{0.573}}   &0.414  &0.064  & \textcolor{blue}{\textbf{0.564}} &0.424  & 0.058\\
      \hline
      \hline
$0.1$ & 0.2 &DRL ($\beta=0$)  &0.39  &0.49  &0.384     & 0.404  &0.477  & 0.367 & 0.427 &0.457  &0.338 \\
& &DRL ($\beta = 0.1$)  &\textcolor{blue}{\textbf{ 0.276}}  &0.595   & 0.752     &\textcolor{blue}{\textbf{ 0.28}} &0.59  & 0.746 & \textcolor{blue}{\textbf{ 0.293}}&0.579  &0.729  \\
& &DRL ($\beta = 0.15$) &\textcolor{blue}{\textbf{ 0.276}}  &0.595   & 0.752   &\textcolor{blue}{\textbf{ 0.28}} &0.59  & 0.746 & \textcolor{blue}{\textbf{0.293}}&0.579  &0.729 \\
& &DRL ($\beta = 0.25$) &\textcolor{blue}{\textbf{0.276}}  &0.595   & 0.752   &\textcolor{blue}{\textbf{ 0.28}} &0.59  & 0.746  & \textcolor{blue}{\textbf{ 0.293}}&0.579  &0.729\\
\cline{2-12}
      &0.3 &DRL ($\beta = 0$)  &0.426  & 0.495  & 0.381     & 0.439 & 0.481 &0.362  & 0.443 &0.478 &0.357 \\
      &&DRL ($\beta = 0.1$)&\textcolor{blue}{\textbf{ 0.376}}  & 0.695 &0.621    &\textcolor{blue}{\textbf{0.387}}   &0.683  & 0.605 & \textcolor{blue}{\textbf{0.388}} &0.601  &0.603\\
       &&DRL ($\beta = 0.15$) &\textcolor{blue}{\textbf{0.376}}  & 0.695 &0.621   &\textcolor{blue}{\textbf{0.387}}   &0.683  & 0.605  &  \textcolor{blue}{\textbf{0.388}} &0.601  &0.603 \\
        &&DRL ($\beta = 0.25$)&\textcolor{blue}{\textbf{0.376}}  & 0.695 &0.621    &\textcolor{blue}{\textbf{0.387}}   &0.683  & 0.605   &  \textcolor{blue}{\textbf{0.388}} &0.601  &0.603 \\
        \cline{2-12}
      &0.4 & DRL ($\beta = 0$)&0.466  & 0.492  &0.362  &  0.489 &0.468  & 0.319 &0.522  &0.435  &0.267\\
 &&DRL ($\beta = 0.1$)& \textcolor{blue}{\textbf{0.465}} & 0.496 & 0.368  & \textcolor{blue}{\textbf{ 0.474}}  &0.485  & 0.353 &\textcolor{blue}{\textbf{ 0.493}}  & 0.46 & 0.321\\
&&DRL ($\beta = 0.15$) & \textcolor{blue}{\textbf{ 0.465}} & 0.496 & 0.368  &\textcolor{blue}{\textbf{ 0.474}}   &0.485  & 0.353 &\textcolor{blue}{\textbf{ 0.493}}  & 0.46 & 0.321  \\
 &&DRL ($\beta = 0.25$)& \textcolor{blue}{\textbf{ 0.465}} & 0.496 & 0.368   & \textcolor{blue}{\textbf{ 0.474}}  &0.485  & 0.353  & \textcolor{blue}{\textbf{ 0.493}}  & 0.463 & 0.321\\
      \hline
      \hline
$0.2$ & 0.2 &DRL ($\beta = 0$)  &0.359  &0.508  &  0.453 & 0.359 &0.508 &0.453 & 0.359 &0.508 &0.453  \\
&& DRL ($\beta = 0.1$) &\textcolor{blue}{\textbf{0.229}}  &0.904  &0.895  &\textcolor{blue}{\textbf{ 0.229}}  &0.904  &0.895 &\textcolor{blue}{\textbf{0.229}}  &0.904  &0.895\\
& & DRL ($\beta = 0.15$) &\textcolor{blue}{\textbf{ 0.229}}  &0.904  &0.895  &\textcolor{blue}{\textbf{0.229}}  &0.904  &0.895 &\textcolor{blue}{\textbf{0.229}}  &0.904  &0.895\\
& &DRL ($\beta = 0.25$) &\textcolor{blue}{\textbf{ 0.229}}  &0.904  &0.895 &\textcolor{blue}{\textbf{0.229}}  &0.904  &0.895  &\textcolor{blue}{\textbf{ 0.229}}  &0.904  &0.895\\
\cline{2-12}
       &0.3 &DRL ($\beta = 0$)& 0.421 & 0.498 &0.454  & 0.421 & 0.498 &0.454 & 0.421 & 0.498 &0.454\\
       &&DRL ($\beta = 0.1$)& \textcolor{blue}{\textbf{0.414}} &0.5  &0.427     & \textcolor{blue}{\textbf{0.414}} &0.499  &0.427  & \textcolor{blue}{\textbf{0.414}} &0.499  &0.427\\
       &&DRL ($\beta = 0.15$) & \textcolor{blue}{\textbf{0.414}} &0.5  &0.427   & \textcolor{blue}{\textbf{0.414}} &0.499  &0.427 & \textcolor{blue}{\textbf{0.414}} &0.499  &0.427\\
       &&DRL ($\beta = 0.25$)& \textcolor{blue}{\textbf{0.414}} &0.5  &0.427    & \textcolor{blue}{\textbf{0.414}} &0.5  &0.427  & \textcolor{blue}{\textbf{ 0.414}} &0.5  &0.427 \\
       \cline{2-12}
      &0.4 &DRL ($\beta = 0$)&0.477  &0.47  &0.4    & 0.486  & 0.459 &0.386  & 0.507  &0.436  &0.351\\
 &&DRL ($\beta = 0.1$)&\textcolor{blue}{\textbf{0.465}}  &0.484  &  0.434 &\textcolor{blue}{\textbf{ 0.465}}  &0.484  &  0.434  &\textcolor{blue}{\textbf{ 0.465}}  &0.483  &  0.433 \\
 &&DRL ($\beta = 0.15$)&\textcolor{blue}{\textbf{ 0.465}}  &0.484  &  0.434 &\textcolor{blue}{\textbf{ 0.465}}  &0.484  &  0.434 &\textcolor{blue}{\textbf{ 0.465}}  &0.483  &  0.434 \\
&&DRL ($\beta = 0.25$)&\textcolor{blue}{\textbf{0.465}}  &0.483  &  0.433   &\textcolor{blue}{\textbf{ 0.465}}  &0.483  &  0.433  &\textcolor{blue}{\textbf{0.465}}  &0.483  &  0.433 \\
\bottomrule
\end{tabular}}
\caption{Results with linear reject option classifier with/without shift trained using Double Ramp Loss ($\mu = 0.95$).}
    \label{Synthetic Data table-DRL}
\end{center}
\end{table*}

\section{Conclusion and Future Work} \label{Conclusion & FW}
In this paper, we give a complete characterization of surrogates calibrated to $\ell_{d}^{\gamma}$ and provide insights on designing them (via extensive analysis of $\ell_{\textrm{ds}}^{\mu, \beta}$ and $\ell_{\textrm{dr}}^{\mu, \beta}$) for the hypothesis set $\mathcal{H}=\mathcal{H}_{\textrm{lin}}$. To the best of our knowledge, this is the first attempt towards analyzing surrogates in the \enquote{Adversarial Robust Reject Option} setting for binary classification from the lens of Calibration Theory. The first line of future work is to provide a proof technique for class of surrogates which are ($\ell_{d}^{\gamma},\mathcal{H}$)-calibrated using the ideas presented in this work for $\ell_{\textrm{ds}}^{\mu, \beta}$ and $\ell_{\textrm{dr}}^{\mu, \beta}$.
Calibration analysis for other function classes like generalized linear models ($\mathcal{H}_{g}$) and single-layer ReLU neural networks ($\mathcal{H}_{\textrm{NN}})$ is another future research direction. 

\bibliography{acml24}

\begin{thebibliography}{28}
\providecommand{\natexlab}[1]{#1}
\providecommand{\url}[1]{\texttt{#1}}
\expandafter\ifx\csname urlstyle\endcsname\relax
  \providecommand{\doi}[1]{doi: #1}\else
  \providecommand{\doi}{doi: \begingroup \urlstyle{rm}\Url}\fi

\bibitem[Awasthi et~al.(2021{\natexlab{a}})Awasthi, Frank, Mao, Mohri, and Zhong]{Awasthi_Calibrn_Consist_adv_sl}
Pranjal Awasthi, Natalie Frank, Anqi Mao, Mehryar Mohri, and Yutao Zhong.
\newblock Calibration and consistency of adversarial surrogate losses.
\newblock In \emph{NeurIPS}, volume~34, 2021{\natexlab{a}}.

\bibitem[Awasthi et~al.(2021{\natexlab{b}})Awasthi, Mao, Mohri, and Zhong]{Awasthi_Finer_calibrn_analysis}
Pranjal Awasthi, Anqi Mao, Mehryar Mohri, and Yutao Zhong.
\newblock A finer calibration analysis for adversarial robustness.
\newblock \emph{CoRR}, abs/2105.01550, 2021{\natexlab{b}}.

\bibitem[Bao et~al.(2020)Bao, Scott, and Sugiyama]{Bao_CSL_Adv_Rob}
Han Bao, Clay Scott, and Masashi Sugiyama.
\newblock Calibrated surrogate losses for adversarially robust classification.
\newblock In \emph{COLT}, volume 125, pages 408--451, 2020.

\bibitem[Bartlett et~al.(2006)Bartlett, Jordan, and McAuliffe]{Bartlett_Jordan_Conv_Clssfn_RB}
Peter Bartlett, Michael Jordan, and Jon McAuliffe.
\newblock Convexity, classification, and risk bounds.
\newblock \emph{Journal of the American Statistical Association}, 101:\penalty0 138--156, 2006.

\bibitem[Bartlett and Wegkamp(2008)]{Bartlett_Hinge_Loss_Reject}
Peter~L. Bartlett and Marten~H. Wegkamp.
\newblock Classification with a reject option using a hinge loss.
\newblock \emph{JMLR}, 9\penalty0 (59):\penalty0 1823--1840, 2008.

\bibitem[Cao et~al.(2022)Cao, Cai, Feng, Gu, GU, An, Niu, and Sugiyama]{NEURIPS2022_03a90e1b}
Yuzhou Cao, Tianchi Cai, Lei Feng, Lihong Gu, Jinjie GU, Bo~An, Gang Niu, and Masashi Sugiyama.
\newblock Generalizing consistent multi-class classification with rejection to be compatible with arbitrary losses.
\newblock In \emph{NeurIPS}, volume~35, pages 521--534, 2022.

\bibitem[Carlini and Wagner(2017)]{Carlini_Wagner_Eval_Rob_NN}
N.~Carlini and D.~Wagner.
\newblock Towards evaluating the robustness of neural networks.
\newblock In \emph{IEEE Symposium on Security and Privacy (SP)}, pages 39--57, 2017.

\bibitem[Chen et~al.(2023)Chen, Raghuram, Choi, Wu, Liang, and Jha]{Chen_Jha_Stratified_Adv_Rob_Rejection}
Jiefeng Chen, Jayaram Raghuram, Jihye Choi, Xi~Wu, Yingyu Liang, and Somesh Jha.
\newblock Stratified adversarial robustness with rejection.
\newblock In \emph{ICML}, pages 4867--4894. PMLR, 2023.

\bibitem[Chow(1970)]{Chow_Optimum_Rec_err_Rej_Trd}
C.~Chow.
\newblock On optimum recognition error and reject tradeoff.
\newblock \emph{IEEE Transactions on Information Theory}, 16\penalty0 (1):\penalty0 41--46, 1970.

\bibitem[Cohen et~al.(2019)Cohen, Rosenfeld, and Kolter]{Cohen_certified_adv_rob_smoothing}
Jeremy Cohen, Elan Rosenfeld, and Zico Kolter.
\newblock Certified adversarial robustness via randomized smoothing.
\newblock In \emph{Proceedings of the 36th International Conference on Machine Learning}, 2019.

\bibitem[Cortes et~al.(2016)Cortes, DeSalvo, and Mohri]{Cortes_Learning_with_Rejn}
Corinna Cortes, Giulia DeSalvo, and Mehryar Mohri.
\newblock Learning with rejection.
\newblock In \emph{COLT}, 2016.

\bibitem[Goodfellow et~al.(2014)Goodfellow, Shlens, and Szegedy]{Goodfellow_Explain_Adv_attk}
Ian Goodfellow, Jonathon Shlens, and Christian Szegedy.
\newblock Explaining and harnessing adversarial examples.
\newblock \emph{arXiv 1412.6572}, 2014.

\bibitem[Kalra et~al.(2021)Kalra, Shah, and Manwani]{DBLP:conf/uai/KalraSM21}
Bhavya Kalra, Kulin Shah, and Naresh Manwani.
\newblock {RISAN:} robust instance specific deep abstention network.
\newblock In \emph{UAI}, volume 161, pages 1525--1534, 2021.

\bibitem[Kato et~al.(2020)Kato, Cui, and Fukuhara]{ATRO}
Masahiro Kato, Zhenghang Cui, and Yoshihiro Fukuhara.
\newblock {ATRO:} adversarial training with a rejection option.
\newblock \emph{CoRR}, abs/2010.12905, 2020.

\bibitem[Madry et~al.(2018)Madry, Makelov, Schmidt, Tsipras, and Vladu]{Madry_DL_Resist_Adv_Attk}
Aleksander Madry, Aleksandar Makelov, Ludwig Schmidt, Dimitris Tsipras, and Adrian Vladu.
\newblock Towards deep learning models resistant to adversarial attacks.
\newblock In \emph{ICLR}, 2018.

\bibitem[Manwani et~al.(2013)Manwani, Desai, Sasidharan, and Sundararajan]{ManwaniDoubleRL}
Naresh Manwani, Kalpit Desai, Sanand Sasidharan, and Ramasubramanian Sundararajan.
\newblock Double ramp loss based reject option classifier.
\newblock In \emph{PAKDD}, 2013.

\bibitem[Meunier et~al.(2022)Meunier, Ettedgui, Pinot, Chevaleyre, and Atif]{Meunier_Consistency_Adv_Classfn}
Laurent Meunier, Raphael Ettedgui, Rafael Pinot, Yann Chevaleyre, and Jamal Atif.
\newblock Towards consistency in adversarial classification.
\newblock In \emph{NeurIPS}, volume~35, pages 8538--8549, 2022.

\bibitem[Ni et~al.(2019)Ni, Charoenphakdee, Honda, and Sugiyama]{Chenri_Calibrn_Multi_Reject}
Chenri Ni, Nontawat Charoenphakdee, Junya Honda, and Masashi Sugiyama.
\newblock On the calibration of multiclass classification with rejection.
\newblock In \emph{NeurIPS}, volume~32, 2019.

\bibitem[Papernot and McDaniel(2018)]{Deep_k_NN_Robust}
Nicolas Papernot and Patrick~D. McDaniel.
\newblock Deep k-nearest neighbors: Towards confident, interpretable and robust deep learning.
\newblock \emph{CoRR}, abs/1803.04765, 2018.

\bibitem[Raghunathan et~al.(2018)Raghunathan, Steinhardt, and Liang]{Aditi_Certified_Defense_Adv_Ex}
Aditi Raghunathan, Jacob Steinhardt, and Percy Liang.
\newblock Certified defenses against adversarial examples.
\newblock \emph{CoRR}, abs/1801.09344, 2018.

\bibitem[Ramaswamy et~al.(2018)Ramaswamy, Tewari, and Agarwal]{Ramaswamy_Consistent_Multi_Reject}
H.~G. Ramaswamy, Ambuj Tewari, and Shivani Agarwal.
\newblock Consistent algorithms for multiclass classification with an abstain option.
\newblock \emph{Electronic Journal of Statistics}, 12:\penalty0 530--554, 2018.

\bibitem[Shah and Manwani(2018)]{Shah_Manwani_Rob_Rej_Lin_Prog}
Kulin Shah and Naresh Manwani.
\newblock Sparse and robust reject option classifier using successive linear programming.
\newblock \emph{CoRR}, abs/1802.04235, 2018.

\bibitem[Shah and Manwani(2020)]{DBLP:conf/aaai/ShahM20}
Kulin Shah and Naresh Manwani.
\newblock Online active learning of reject option classifiers.
\newblock In \emph{AAAI}, pages 5652--5659. {AAAI} Press, 2020.

\bibitem[Steinwart(2007)]{Steinwart_Compare_Diff_Losses}
Ingo Steinwart.
\newblock How to compare different loss functions and their risks.
\newblock \emph{Constructive Approximation}, 26:\penalty0 225--287, 08 2007.

\bibitem[Szegedy et~al.(2013)Szegedy, Zaremba, Sutskever, Bruna, Erhan, Goodfellow, and Fergus]{Szegedy_Intrigue_prop_NN}
Christian Szegedy, Wojciech Zaremba, Ilya Sutskever, Joan Bruna, D.~Erhan, Ian~J. Goodfellow, and Rob Fergus.
\newblock Intriguing properties of neural networks.
\newblock \emph{CoRR}, abs/1312.6199, 2013.

\bibitem[Wong and Kolter(2018)]{Wong_Kolter_Adv_Ex_Cvx}
Eric Wong and Zico Kolter.
\newblock Provable defenses against adversarial examples via the convex outer adversarial polytope.
\newblock In \emph{ICML}, pages 5286--5295, 2018.

\bibitem[Yang et~al.(2019)Yang, Chen, Hsieh, Wang, and Jordan]{ML_LOO_Adv_exmp_Feature_Attrn}
Puyudi Yang, Jianbo Chen, Cho{-}Jui Hsieh, Jane{-}Ling Wang, and Michael~I. Jordan.
\newblock {ML-LOO:} detecting adversarial examples with feature attribution.
\newblock \emph{CoRR}, abs/1906.03499, 2019.

\bibitem[Zhang et~al.(2022)Zhang, Jiang, He, and Wang]{Certified_rob_Lipschtiz_NN_Boolean}
Bohang Zhang, Du~Jiang, Di~He, and Liwei Wang.
\newblock Rethinking lipschitz neural networks and certified robustness: A boolean function perspective, 2022.

\end{thebibliography}

\appendix
\section{Proof of Proposition~\ref{prop1}} 
\label{app1}


\begin{proof}
Let $\ell$ be a non-increasing function of $yf(\mathbf{x})$. The following property holds for $\ell$.
\begin{equation}\label{non-inc-property}
    \sup_{\mathbf{x}} \; \ell(yf(\mathbf{x})) = \ell (\inf_{\mathbf{x}} \; yf(\mathbf{x}))
\end{equation}
Both indicator functions, $\mathbbm{1}_{ \{ y f\left(\mathbf{x}^{\prime} \right ) < -\rho\}}$ and $\mathbbm{1}_{ \{ y f\left(\mathbf{x}^{\prime} \right ) <\rho\}}$, are non-increasing with $y f\left(\mathbf{x}^{\prime} \right )$.  Hence using ($\ref{non-inc-property}$) in the definition of the Adversarial Robust Reject Option Loss, 
we have 
\begin{equation}\label{l_gamma_d_defn}
    \ell^{\gamma}_{d}(yf(\mathbf{x}),\rho) = (1-d) \; \mathbbm{1}_{ \{ \; \inf_{\mathbf{x}^{\prime}:\left\|\mathbf{x}-\mathbf{x}^{\prime}\right\| \leq \gamma} y f\left(\mathbf{x}^{\prime} \right ) < -\rho \; \}} + \; d \; \mathbbm{1}_{ \{ \; \inf_{\mathbf{x}^{\prime}:\left\|\mathbf{x}-\mathbf{x}^{\prime}\right\| \leq \gamma} y f\left(\mathbf{x}^{\prime} \right ) \leq \rho \; \}}.
\end{equation}

For $\mathcal{H}_{\textrm{lin}}$, $f(\mathbf{x}) = \mathbf{w} \cdot \mathbf{x}~$ with $~\| \mathbf{w} \| = 1$.
The optimization problem formulated in eq.~($\ref{l_gamma_d_defn}$) is as follows :

\begin{equation} \label{h_lin_opt}
\begin{aligned}
\min_{\mathbf{x}^{\prime}} \quad & y \;( \mathbf{w} \cdot \mathbf{x}^{\prime})\\
\textrm{s.t.} \quad &  \| \mathbf{x} - \mathbf{x}^{\prime} \| \leq \gamma    \\
\end{aligned}
\end{equation}

($\ref{h_lin_opt}$) is a convex optimization problem. The Lagrangian is given by 
\begin{align*}
    \mathcal{L}(\mathbf{x}^{\prime} , \lambda) = y~\mathbf{
    w} \cdot \mathbf{x}^{\prime} + \lambda~(\| \mathbf{x} - \mathbf{x}^{\prime} \| - \gamma) 
\end{align*}

where $\lambda \in \mathbb{R}$ is a Lagrangian multiplier. Applying KKT conditions, we get the following
\begin{enumerate}
    \item $y~\mathbf{w} - \lambda\;\frac{\mathbf{x} - \mathbf{x}^{\prime}}{\| \mathbf{x} - \mathbf{x}^{\prime}\|} = 0$

    \item $\lambda \geq 0$
    
    \item $\| \mathbf{x} - \mathbf{x}^{\prime}\| \leq \gamma$

    \item $\lambda \; (\| \mathbf{x} - \mathbf{x}^{\prime}\| - \gamma) = 0$ \quad (Complementary Slackness)
\end{enumerate}
 Using the condition from complementary slackness , we have the trivial case when $\lambda = 0$ as the objective function value is always $0$. For $\lambda \neq 0$, it holds that $\| \mathbf{x} - \mathbf{x}^{\prime}\| = \gamma$. Hence, the constraint $\| \mathbf{x} - \mathbf{x}^{\prime}\| \leq \gamma$ is activated. From 1.) we have $\mathbf{x} - \mathbf{x}^{\prime} = \frac{y \; \gamma}{\lambda} \mathbf{w}$. But, $\| \mathbf{x} - \mathbf{x}^{\prime}\| = \gamma$, so $\| \frac{y \; \gamma}{\lambda}  \mathbf{w} \| = \gamma$. Solving, we get $\lambda = \| \mathbf{w} \|$ and $\mathbf{x} - \mathbf{x}^{\prime} = \frac{y\;\gamma}{\|\mathbf{w}\|} \mathbf{w}$.

The optimal solution to (\ref{h_lin_opt}) is given by $(\mathbf{x}^{\prime})^{*} = \mathbf{x} - \frac{y\;\gamma\;\mathbf{w}}{\| \mathbf{w} \|}$. Substituting this in (\ref{l_gamma_d_defn}), we get 

\begin{equation} 
       \ell^{\gamma}_{d}(yf(\mathbf{x}),\rho) = (1-d) \; \mathbbm{1}_{ \{  y f\left(\mathbf{x} \right ) < -\rho + \gamma \; \}} + \; d \; \mathbbm{1}_{ \{ \;  y f\left(\mathbf{x} \right ) \leq \;\rho +\gamma \; \}}
\end{equation}
which is equivalent to $\gamma$-right shift of $\ell_{d}$.

\end{proof}

\section{Proof of Lemma~\ref{lemma:excess-risk-target}}
\label{sec:proof-excess-risk-target}

\begin{proof}
A case by case breakdown of the definition of $\mathcal{C}_{\ell^{\gamma}_{d} , \mathcal{H}}(\alpha, \eta)$ is the first step. 
Depending upon prediction or rejection, the minimal inner-risk will be   $\mathcal{C}_{\ell^{\gamma}_{d} , \mathcal{H}}^{*}(\eta) = \mathcal{C}_{\ell^{\gamma}_{d} , \mathcal{H}}^{*}(\alpha, \eta) = \min \{ \eta , 1-\eta , d\}$. Additionally, we also assume that $\rho + \gamma < 1$. For each sub-case, a further splitting is done based on the minimal inner-risk value and using $\Delta \mathcal{C}_{\ell^{\gamma}_{d} , \mathcal{H}}(\alpha, \eta)$, 
we have the desired result. We prove it for one of the cases and for rest of the cases, proof follows the similar procedure. Consider the case of $\alpha < -\rho -\gamma$ : when the minimal inner risk is $d$, then, 
$\Delta \mathcal{C}_{\ell^{\gamma}_{d} , \mathcal{H}}(\alpha, \eta) = \mathcal{C}_{\ell^{\gamma}_{d} , \mathcal{H}}(\alpha, \eta) - \Delta \mathcal{C}_{\ell^{\gamma}_{d} , \mathcal{H}}^{*}(\alpha, \eta) = \eta - d$. When the minimal inner risk is $\eta$, $\Delta \mathcal{C}_{\ell^{\gamma}_{d} , \mathcal{H}}(\alpha, \eta) = \eta - \eta = 0$ and for the sub-case when minimal inner risk is $1-\eta$, $\Delta \mathcal{C}_{\ell^{\gamma}_{d} , \mathcal{H}}(\alpha, \eta) = \eta - (1-\eta) = 2\eta - 1$. Combining all the sub-cases using indicator functions, we have for 
\begin{equation*}
    \Delta \mathcal{C}_{\ell^{\gamma}_{d} , \mathcal{H}}(\alpha, \eta) = (\eta - d)\; \mathbbm{1}_{\min \{\eta,1-\eta \} - d \geq 0} + | 2\eta -1 |\;\mathbbm{1}_{2\eta -1 >0}\;\mathbbm{1}_{\min \{\eta,1-\eta\} - d < 0}
\end{equation*}
\end{proof}

\section{Proof of Theorem~\ref{thm1}} 
\label{calibrn_theorem_proof}


\begin{proof}
Let $\ell$ be a margin-based surrogate to $\ell_{d}^{\gamma}$. Using Proposition 5, 
we have that  $\ell$ is ($\ell_{d}^{\gamma} , \mathcal{H}$)-calibrated if and only if its corresponding calibration function $\delta(\epsilon) > 0 , \;\forall \epsilon$. The case for $\eta = 0.5$ is dealt separately. Based on range of $\eta$, two cases are made and for each one, a further split is made based on prediction or rejection, and then, the calibration function is computed. This further has 3 sub-cases - (based on the \enquote{Bayes classifier} and change in definition of $C_{\ell_{d}^{\gamma} , \mathcal{H}}$) 
\begin{enumerate}
    \item $1-\eta < d$ 
    \item $d \leq 1-\eta \;$ and $\;\eta \geq \eta_{\text{right}}$
    \item $d \leq 1-\eta \;$ and $\; \eta < \eta_{\text{right}}$
\end{enumerate}

\begin{equation} 
\delta(\epsilon) = \delta_{1}  +  \delta_{2} 
\end{equation} 

where 
\begin{equation} \delta_{1} = \delta_{r}(\epsilon)\;\mathbbm{1}_{\{ \min (\eta,1-\eta) -d \geq 0\}}
\end{equation} 

\begin{equation}
\delta_{2} = \delta_{p}(\epsilon)\;\mathbbm{1}_{\{ \min (\eta,1-\eta) -d < 0\}} 
\end{equation} 


For $\delta(\epsilon) > 0 \;$ to hold  $\forall \eta \in  \left[ 0,1 \right] $ , we need either $\delta_{1}(\epsilon) > 0$ or $\delta_{2}(\epsilon) > 0\;$ to hold $\forall \eta \in  \left[ 0,1 \right] $. Note that one among $\delta_{1}$ or $\delta_{2}$ is always $0$. We use Figure~$\ref{excess_target_risk vs eta-plot}$ to split the case of $\eta > 0.5$, into further sub-cases. First split is based on prediction or rejection, i.e minimizer is $\{ \eta, 1-\eta\}$ or $d$. In the prediction case, a definition change occurs around the points $\eta_{\textrm{left}}$ (when $\eta < 0.5$) or $\eta_{\textrm{right}}$ (when $\eta > 0.5)$, as seen in Figure~$\ref{excess_target_risk vs eta-plot}$. \\

Case i) : $\eta > \frac{1}{2}$

Sub-case A) : $\min \{ \eta,1-\eta \} < d$ (prediction)

\begin{equation} \label{delta_case1}
\delta_{p}(\epsilon) =
    \begin{cases}
        \infty & \text{if } \epsilon > \eta - (1-\eta)(1-d)\\
        
        \inf_{\{ \alpha : -\rho -\gamma \leq \alpha < -\rho + \gamma \} } \Delta \mathcal{C}_{\ell , \mathcal{H}} (\alpha,\eta)  & \text{if }  \eta - (1-\eta)(1-d) \geq \epsilon > 2\eta - 1 \\
        
        \inf_{ \{ \alpha : \alpha < -\rho - \gamma \} } \Delta \mathcal{C}_{\ell , \mathcal{H}} (\alpha,\eta) & \text{if } 2\eta - 1 \geq \epsilon > \eta \; d\\
        
        \inf_{ \{ \alpha : \rho -\gamma < \alpha \leq  \rho  + \gamma \} } \Delta \mathcal{C}_{\ell , \mathcal{H}} (\alpha,\eta) & \text{if } \eta\;d \geq \epsilon > d - (1-\eta)  \\

        \inf_{ \{ \alpha : -\rho + \gamma \leq \alpha \leq \rho - \gamma \} } \Delta \mathcal{C}_{\ell , \mathcal{H}} (\alpha,\eta) & \text{if }  d - (1-\eta) \geq \epsilon \\
        
    \end{cases}
\end{equation}

Sub-case B) : $\min \{ \eta,1-\eta \} \geq d$ (rejection)

I) $\eta \geq \eta_{\text{right}}$

\begin{equation} \label{delta_case2}
\delta_{r}(\epsilon) =
    \begin{cases}
        \infty & \text{if } \epsilon > (1-d)\;\eta \\
        
        \inf_{\{ \alpha : -\rho -\gamma \leq \alpha < -\rho + \gamma \} } \Delta \mathcal{C}_{\ell , \mathcal{H}} (\alpha,\eta)  & \text{if }  (1-d)\;\eta \geq \epsilon > \eta - d \\
        
        \inf_{ \{ \alpha : \alpha < -\rho - \gamma \} } \Delta \mathcal{C}_{\ell , \mathcal{H}} (\alpha,\eta) & \text{if } \eta - d \geq \epsilon > (1-\eta)(1-d)\\
        
        \inf_{ \{ \alpha : \rho -\gamma < \alpha \leq  \rho  + \gamma \} } \Delta \mathcal{C}_{\ell , \mathcal{H}} (\alpha,\eta) & \text{if } (1-\eta)(1-d) \geq \epsilon > 1-\eta-d  \\

        \inf_{ \{ \alpha : -\rho + \gamma \leq \alpha \leq \rho - \gamma \} } \Delta \mathcal{C}_{\ell , \mathcal{H}} (\alpha,\eta) & \text{if }  1-\eta-d \geq \epsilon \\
        
    \end{cases}
\end{equation}

II) $\eta < \eta_{\text{right}}$ [Narrow band]

\begin{equation} \label{delta_case3}
\delta_{r}(\epsilon) =
    \begin{cases}
        \infty & \text{if } \epsilon > (1-d)\;\eta \\
        
        \inf_{\{ \alpha : -\rho -\gamma \leq \alpha < -\rho + \gamma \} } \Delta \mathcal{C}_{\ell , \mathcal{H}} (\alpha,\eta)  & \text{if }  (1-d)\;\eta \geq \epsilon > (1-\eta)(1-d) \\
        
        \inf_{ \{ \alpha : \alpha < -\rho - \gamma \} } \Delta \mathcal{C}_{\ell , \mathcal{H}} (\alpha,\eta) & \text{if } (1-\eta)(1-d) \geq \epsilon > \eta - d\\
        
        \inf_{ \{ \alpha : \rho -\gamma < \alpha \leq  \rho  + \gamma \} } \Delta \mathcal{C}_{\ell , \mathcal{H}} (\alpha,\eta) & \text{if } \eta - d \geq \epsilon > 1-\eta-d  \\

        \inf_{ \{ \alpha : -\rho + \gamma \leq \alpha \leq \rho - \gamma \} } \Delta \mathcal{C}_{\ell , \mathcal{H}} (\alpha,\eta) & \text{if }  1-\eta-d \geq \epsilon \\
        
    \end{cases}
\end{equation}

\textbf{NOTE:} For margin-based surrogate, $\mathcal{C}_{\ell,\mathcal{H}}(f(\mathbf{x}), \eta)$ and $\Delta \mathcal{C}_{\ell , \mathcal{H}}(f(\mathbf{x}), \eta)$ are symmetrical about $\eta = \frac{1}{2}$. Hence, the definitions for Case ii) can be obtained by replacing $\eta$ with $1-\eta$ from Case i).

Case ii) : $\eta < \frac{1}{2}$ \\

Sub-case A) : $\min \{ \eta,1-\eta \} < d$ (prediction)

\begin{equation}
\delta_{p}(\epsilon) =
    \begin{cases}
        \infty & \text{if } \epsilon > (1-\eta) - \eta(1-d)\\
        
        \inf_{\{ \alpha : -\rho -\gamma \leq \alpha < -\rho + \gamma \} } \Delta \mathcal{C}_{\ell , \mathcal{H}} (\alpha,\eta)  & \text{if }  (1-\eta) - \eta(1-d) \geq \epsilon > 1-2\eta \\
        
        \inf_{ \{ \alpha : \alpha < -\rho - \gamma \} } \Delta \mathcal{C}_{\ell , \mathcal{H}} (\alpha,\eta) & \text{if } 1-2\eta \geq \epsilon > (1-\eta)\; d\\
        
        \inf_{ \{ \alpha : \rho -\gamma < \alpha \leq  \rho  + \gamma \} } \Delta \mathcal{C}_{\ell , \mathcal{H}} (\alpha,\eta) & \text{if } (1-\eta)\; d \geq \epsilon > d - \eta  \\

        \inf_{ \{ \alpha : -\rho + \gamma \leq \alpha \leq \rho - \gamma \} } \Delta \mathcal{C}_{\ell , \mathcal{H}} (\alpha,\eta) & \text{if }  d - \eta \geq \epsilon \\
        
    \end{cases}
\end{equation}

Sub-case B) : $\min \{ \eta,1-\eta \} \geq d$ (rejection)

I) $\eta \leq \eta_{\text{left}}$

\begin{equation}
\delta_{r}(\epsilon) =
    \begin{cases}
        \infty & \text{if } \epsilon > (1-d)\;(1-\eta) \\
        
        \inf_{\{ \alpha : -\rho -\gamma \leq \alpha < -\rho + \gamma \} } \Delta \mathcal{C}_{\ell , \mathcal{H}} (\alpha,\eta)  & \text{if }  (1-d)\;(1-\eta) \geq \epsilon > 1-\eta - d \\
        
        \inf_{ \{ \alpha : \alpha < -\rho - \gamma \} } \Delta \mathcal{C}_{\ell , \mathcal{H}} (\alpha,\eta) & \text{if } 1-\eta - d \geq \epsilon > \eta\;(1-d)\\
        
        \inf_{ \{ \alpha : \rho -\gamma < \alpha \leq  \rho  + \gamma \} } \Delta \mathcal{C}_{\ell , \mathcal{H}} (\alpha,\eta) & \text{if } \eta\;(1-d) \geq \epsilon > \eta-d  \\

        \inf_{ \{ \alpha : -\rho + \gamma \leq \alpha \leq \rho - \gamma \} } \Delta \mathcal{C}_{\ell , \mathcal{H}} (\alpha,\eta) & \text{if }  \eta-d \geq \epsilon \\
        
    \end{cases}
\end{equation}

II) $\eta > \eta_{\text{left}}$ [Narrow band]

\begin{equation}
\delta_{r}(\epsilon) =
    \begin{cases}
        \infty & \text{if } \epsilon > (1-d)\;(1-\eta) \\
        
        \inf_{\{ \alpha : -\rho -\gamma \leq \alpha < -\rho + \gamma \} } \Delta \mathcal{C}_{\ell , \mathcal{H}} (\alpha,\eta)  & \text{if }  (1-d)\;(1-\eta) \geq \epsilon > (1-d)\;\eta \\
        
        \inf_{ \{ \alpha : \alpha < -\rho - \gamma \} } \Delta \mathcal{C}_{\ell , \mathcal{H}} (\alpha,\eta) & \text{if } (1-d)\;\eta  \geq \epsilon > 1-\eta-d\\
        
        \inf_{ \{ \alpha : \rho -\gamma < \alpha \leq  \rho  + \gamma \} } \Delta \mathcal{C}_{\ell , \mathcal{H}} (\alpha,\eta) & \text{if } 1-\eta-d \geq \epsilon > \eta-d  \\
        
        \inf_{ \{ \alpha : -\rho + \gamma \leq \alpha \leq \rho - \gamma \} } \Delta \mathcal{C}_{\ell , \mathcal{H}} (\alpha,\eta) & \text{if }  \eta-d \geq \epsilon \\
        
    \end{cases}
\end{equation}

For each of these sub-cases, we arrive at this definition using the graph below.






For each case, the corresponding calibration function definitions are ($\ref{delta_case1}$) , ($\ref{delta_case2}$) and ($\ref{delta_case3}$) respectively. Using \textbf{Proposition 5}, 
it holds that $\ell$ is ($\ell_{d}^{\gamma} , \mathcal{H}$)-calibrated if and only if its corresponding calibration function $\delta(\epsilon) > 0$. Applying this for each case , we get 

\begin{enumerate} \label{eta_gt_0.5_calibrn_cond}
    \item $\inf_{\alpha > \rho + \gamma} \mathcal{C}_{\ell, \mathcal{H}}(\alpha,1-\eta) > \inf_{\alpha \in \text{Z}} \mathcal{C}_{\ell, \mathcal{H}}(\alpha,\eta)$ 

    $\equiv$

    $\inf_{\alpha < -\rho - \gamma} \mathcal{C}_{\ell, \mathcal{H}}(\alpha,\eta) > \inf_{\alpha \in \text{Z}} \mathcal{C}_{\ell, \mathcal{H}}(\alpha,\eta)$

    \item $\inf_{\rho - \gamma < \alpha 
    \leq \rho + \gamma} \mathcal{C}_{\ell, \mathcal{H}}(\alpha,1-\eta) > \inf_{\alpha \in \text{Z}} \mathcal{C}_{\ell, \mathcal{H}}(\alpha,\eta)$ 

    $\equiv$ 

    $\inf_{-\rho - \gamma \leq \alpha 
    \leq -\rho + \gamma} \mathcal{C}_{\ell, \mathcal{H}}(\alpha,\eta) > \inf_{\alpha \in \text{Z}} \mathcal{C}_{\ell, \mathcal{H}}(\alpha,\eta)$  

    \item $\inf_{| \alpha | \leq \rho - \gamma} \mathcal{C}_{\ell, \mathcal{H}}(\alpha,\eta) > \inf_{\alpha \in \text{Z}} \mathcal{C}_{\ell, \mathcal{H}}(\alpha,\eta)$

    $\equiv$

    $\inf_{-\rho + \gamma \leq \alpha 
    \leq \rho - \gamma} \mathcal{C}_{\ell, \mathcal{H}}(\alpha,\eta) > \inf_{\alpha \in \text{Z}} \mathcal{C}_{\ell, \mathcal{H}}(\alpha,\eta)$ 

    \item $\inf_{\rho - \gamma < \alpha 
    \leq \rho + \gamma} \mathcal{C}_{\ell, \mathcal{H}}(\alpha,\eta) > \inf_{\alpha \in \text{Z}} \mathcal{C}_{\ell, \mathcal{H}}(\alpha,\eta)$ \label{no_cvx_clb}  

\end{enumerate}
where $ \text{Z} = \bigl[ - \| \mathbf{x} \| \;, \; \| \mathbf{x} \| \bigr] $.

By combining all 4 conditions mentioned above, we get 
\begin{equation} 
\inf_{ - \| \mathbf{x} \| \leq \alpha \leq \rho + \gamma} \mathcal{C}_{\ell, \mathcal{H}}(\alpha,\eta) > \inf_{-\| \mathbf{x} \| \alpha \leq \| \mathbf{x} \|} \mathcal{C}_{\ell, \mathcal{H}}(\alpha,\eta)
\end{equation}

Now, we compute $\delta(\epsilon)$ for $\eta = 0.5$\\ 
\begin{equation}
    \Delta \mathcal{C}_{\ell_{d}^{\gamma} , \mathcal{H}}(\alpha,\eta) = 
    \begin{cases}
        0 & \text{if } \; | \alpha | \leq \rho - \gamma \\
        \frac{1-d}{2} & \text{if } \; \rho - \gamma < | \alpha | \leq \rho + \gamma \\
        \frac{1}{2} - d &  \text{if } \; \rho + \gamma < | \alpha |  \\
    \end{cases}
\end{equation} 

\begin{equation} \label{delta_piecewise_eta_0.5}
    \delta (\epsilon) = 
    \begin{cases}
        \infty & \text{if } \; | \alpha | \leq \rho - \gamma \; \text{or } \; \epsilon > \frac{1-d}{2} \\
        
        \inf_{\rho - \gamma < | \alpha | \leq \rho + \gamma} \Delta C_{\ell , \mathcal{H}}\; (\alpha,\frac{1}{2}) & \text{if } \; | \alpha | > \rho - \gamma \; \text{and }\; \frac{1-d}{2} \geq \epsilon > \frac{1}{2} - d\\

         \inf_{ | \alpha | > \rho + \gamma} \Delta C_{\ell , \mathcal{H}}\; (\alpha,\frac{1}{2}) & \text{if } \; | \alpha | > \rho - \gamma \; \text{and }\;  \frac{1}{2} - d \geq \epsilon\\
    \end{cases}
\end{equation} 

\begin{center}
    Bayes-inner risk : $\mathcal{C}^{*}_{\ell^{\gamma}_{d}, \mathcal{H}} (\alpha, \frac{1}{2}) = d$
\end{center}

Using \textbf{Proposition 5} 
on $(\ref{delta_piecewise_eta_0.5})$, we get \\ 

A surrogate $\ell$ is ($\ell^{\gamma}_{d} , \mathcal{H}$) calibrated if and only if
\begin{equation} \label{eta_0.5_calibrn_cond1}
    \inf_{\rho - \gamma < | \alpha | \leq \rho + \gamma} \Delta \mathcal{C}_{\ell,\mathcal{H}} (\alpha, \frac{1}{2}) > 0  \implies \inf_{\rho - \gamma < \alpha \leq \rho + \gamma} \mathcal{C}_{\ell, \mathcal{H}} (\alpha,\frac{1}{2}) > \inf_{\alpha \in \text{Z}} \mathcal{C}_{\ell, \mathcal{H}} (\alpha,\frac{1}{2}) \\
\end{equation}
and 
\begin{equation} \label{eta_0.5_calibrn_cond2}
  \inf_{| \alpha | > \rho + \gamma} \Delta \mathcal{C}_{\ell,\mathcal{H}} (\alpha, \frac{1}{2}) > 0  \implies \inf_{\alpha  > \rho + \gamma} \mathcal{C}_{\ell, \mathcal{H}} (\alpha,\frac{1}{2}) > \inf_{\alpha \in \text{Z}} \mathcal{C}_{\ell, \mathcal{H}} (\alpha,\frac{1}{2})   
\end{equation} \\

By combining ($\ref{eta_0.5_calibrn_cond1}$) and ($\ref{eta_0.5_calibrn_cond2}$) , we get 

\begin{equation} 
   \inf_{\rho - \gamma < \alpha \leq \| \mathbf{x} \| } \mathcal{C}_{\ell, \mathcal{H}} (\alpha,\frac{1}{2}) > \inf_{0 \leq \alpha \leq \| \mathbf{x} \|} \mathcal{C}_{\ell, \mathcal{H}} (\alpha,\frac{1}{2}) 
\end{equation}

\textbf{NOTE :} $\mathcal{C}_{\ell , \mathcal{H}} (\alpha , \frac{1}{2})$ is symmetric about $0$. So, $\mathcal{C}^{*}_{\ell,\mathcal{H}} (\alpha,\frac{1}{2}) = \inf_{0 \leq \alpha \leq \| \mathbf{x} \| }\mathcal{C}_{\ell,\mathcal{H}} (\alpha,\frac{1}{2})$.

Thus, for any surrogate $\ell$ to be ($\ell_{d}^{\gamma} , \mathcal{H}$)-calibrated if and only if it satisfies ($\ref{main_calibrn_cond_1}$) and ($\ref{main_calibrn_cond_2}$). 
 
\end{proof}

\section{Proof of Theorem~\ref{thm:negative-result-convex-surrogates}}
\label{proof:negative-result-convex-surrogates}


\begin{proof}
Assume that $\ell$, convex, differentiable surrogate to $\ell_{d}^{\gamma}$ is $\mathcal{H}$-calibrated. For $\eta = \frac{1}{2}$, the minimizer of the conditional risk lies at $0$. As $\ell$ is convex, $\mathcal{C}_{\ell,\mathcal{H}}(\alpha,\eta)$ is also convex and $\mathcal{C}_{\ell,\mathcal{H}}(\alpha,\frac{1}{2}) = 0.5~ \bar{\ell}(\alpha)$. From convexity of $\bar{\ell}$, we have $\bar{\ell}(0) \leq \bar{\ell}(\alpha)$ $\forall \alpha$. Thus, calibration condition ($\ref{main_calibrn_cond_1}$) is satisfied. Next, we consider the case when $\eta > \frac{1}{2}$ and use proof by contradiction.

Any convex function on a compact set $[\theta_{1},\theta_{2}]  \subset \mathbb{R} $ can be characterised as : 
\begin{enumerate}
    \item Non-increasing 
    \item Non-decreasing
    \item Non-increasing upto $\omega$ ($\in [\theta_{1},\theta_{2}])$ and non-decreasing on $[\omega , \theta_{2}]$
\end{enumerate}

Using the above characterization, for calibration condition ($\ref{main_calibrn_cond_2}$) to hold, two cases are possible.

Let $\alpha^{\star} = \textrm{argmin}_{\alpha}~ \mathcal{C}_{\ell,\mathcal{H}}(\alpha,\eta)$. \\

\textbf{Case i) :} $\rho + \gamma  < \alpha^{\star} \leq \| \mathbf{x} \| $ (when there exists a minimizer inside the compact set) 
\begin{equation*}
 \frac{\mathrm{d}}{\mathrm{d} \alpha} \mathcal{C}_{\ell,\mathcal{H}}(\alpha,\eta)  \bigg \vert_{\alpha = \alpha^{\star}} = 0 
\end{equation*}

\begin{equation*}
  \therefore \;\eta ~ \ell^{'}(\alpha^{\star}) = (1-\eta)~ \ell^{'}(-\alpha^{\star})    
\end{equation*}

As $\eta \in (\frac{1}{2},1]$, we have that $\frac{\eta}{1-\eta} > 1$. Thus, $\ell^{'}(-\alpha^{\star}) > \ell^{'}(\alpha^{\star}) $. But as $\ell$ is convex, $\ell^{'}$ is monotone. Hence, $[\ell^{'}(\alpha^{\star})- \ell^{'}(-\alpha^{\star})] (2~\alpha^{\star}) \geq 0$. This implies that $\ell^{'}(\alpha^{\star}) \geq \ell^{'}(-\alpha^{\star}) $. Hence, we have arrived at a contradiction. \\

\textbf{Case ii) :} $\alpha^{\star} > \| \mathbf{x} \| \;$ (non-increasing on the compact set)

Then, it holds that $\frac{\mathrm{d}}{\mathrm{d} \alpha} \mathcal{C}_{\ell,\mathcal{H}}(\alpha,\eta)  \bigg \vert_{\alpha = \| \mathbf{x} \|} < 0 $ and since ($\ref{main_calibrn_cond_2}$) requires the minima to lie in $(\rho + \gamma, \| \mathbf{x} \|~ ]$, the following must hold :

\begin{equation*}
\mathcal{C}_{\ell,\mathcal{H}} (\rho + \gamma , \eta) > \mathcal{C}_{\ell,\mathcal{H}} (\| \mathbf{x} \|, \eta) \quad \forall \mathbf{x} ~ \textrm{such that} ~ \| \mathbf{x} \| > \rho + \gamma  
\end{equation*}
Using the definition of conditional risk and rearranging the terms, we get 
\begin{equation*}
\bigg( \frac{\eta}{1-\eta} \bigg) [~\ell(\| \mathbf{x} \|) - \ell(\rho + \gamma)~] < [~\ell(-\rho - \gamma) - \ell(-\| \mathbf{x} \|)~]
\end{equation*}

Since $\eta \in (\frac{1}{2},1]$, it holds that $\frac{\eta}{1-\eta} > 1$. Thus, $\ell(\| \mathbf{x} \|) - \ell(\rho + \gamma) < \ell(-\rho - \gamma) - \ell(-\| \mathbf{x} \|) $ which implies that $\bar{\ell} (\| \mathbf{x} \|) < \bar{\ell}(\rho + \gamma)$. But, $\bar{\ell}$ is a even, convex function. Hence, it holds that $\bar{\ell} (\| \mathbf{x} \|) - \bar{\ell} (\rho + \gamma) \geq \bar{\ell}^{'} (\rho + \gamma) ~[~\| \mathbf{x} \| -
(\rho + \gamma)  ~] $ and $\bar{\ell}^{'}(0) = 0$. So, $\bar{\ell}^{'}(\rho + \gamma) > 0$ and we have that $\bar{\ell} (\| \mathbf{x} \|) \geq \bar{\ell}(\rho + \gamma)$ resulting in a contradiction.

Thus, no differentiable convex surrogate is ($\ell^{\gamma}_{d},\mathcal{H}$)-calibrated.
  
\end{proof}

\section{Proof of Theorem~\ref{theorem_neg_calibrn}}
\label{append-neg_calibrn}

\begin{proof}
Let $\ell$ be a margin-based surrogate whose $\mathcal{C}_{\ell , \mathcal{H}}(\alpha , \eta)$ is quasi-concave in $\alpha$ , $\forall \eta \in \left[ 0, 1 \right]$ and assume that $\ell$ is ($\ell_{d}^{\gamma},\mathcal{H}$)-calibrated. Let $\Bar{\ell} (\alpha) = \ell(\alpha) + \ell(-\alpha)$. Since it holds true $\forall \eta \in [0,1]$, it must hold for $\eta = \frac{1}{2}$. At $\eta = \frac{1}{2}$, Quasi-concavity is transferred onto $\Bar{\ell}$. Also, every quasi-concave function on $\mathbb{R}$ can be characterized as following : 

\begin{enumerate}
    \item non-increasing on $\mathbb{R}$ 
    \item non-decreasing on $\mathbb{R}$
    \item non-decreasing up to a point of maxima $\theta$ i.e on $(-\infty , \theta] $, constant upto to $\omega$ ($\theta \leq \omega$) and non-increasing on $[ \omega , \infty )$.
\end{enumerate}

Also, $\Bar{\ell}$ is symmetric about $0$. Hence, Quasi-concavity for even function would imply that first two cases essentially are reduced to constant functions. Else, third case prevails and we get two scenarios, maxima on either side of $0$, both of which imply that $\Bar{\ell}(\alpha)$ is non-increasing for $\alpha > 0$. For any surrogate $\ell$ to be $(\ell_d^{\gamma} , \mathcal{H})$-calibrated , it must satisfy ($\ref{main_calibrn_cond_1}$) i.e

\begin{align*}
    \inf_{\alpha \; \in \; (\rho-\gamma, \| x \|\; ]} \mathcal{C}_{\ell , \mathcal{H}} (\alpha,\frac{1}{2}) > \inf_{\alpha \; \in \; [0, \| x \|\; ]} \mathcal{C}_{\ell , \mathcal{H}} (\alpha,\frac{1}{2}) 
\end{align*}

\begin{align*}
    \therefore ~ \inf_{\alpha \; \in \; (\rho-\gamma, \| x \|\; ]} \Bar{\ell}(\alpha) >  \inf_{\alpha \; \in \; [0, \| x \|\; ]} \Bar{\ell}(\alpha)    
\end{align*}

This is in contradiction to $\Bar{\ell}(\alpha)$ being non-increasing. Hence, our initial assumption was incorrect. \\

\end{proof}

\section{Reproducibility}
Link to the repository containing the code files for reproducing the simulations is given \href{https://github.com/Vrund0212/Calibrated-Losses-for-Adversarial-Robust-Reject-Option-Classification}{here}.


%











\end{document}